%% file: iclr2026_conference.tex
\documentclass{article} %
\usepackage{iclr2026_conference}
\usepackage{times}

\input{math_commands.tex}

\usepackage{hyperref}
\usepackage[capitalise]{cleveref}
\usepackage{amsthm}
\usepackage{url}
\usepackage{graphicx}
\usepackage{listings}
\usepackage{booktabs}
\usepackage{cancel}
\usepackage{xcolor}
\usepackage{mathtools}
\usepackage{wrapfig}
\usepackage{algorithm}
\usepackage{algpseudocode}
\usepackage{longtable}

\usepackage{pifont}
\definecolor{myred}{RGB}{215,48,39}
\definecolor{mygreen}{RGB}{26,152,80}

\definecolor{chipgreen}{HTML}{d4ecd4}
\definecolor{chipgray}{HTML}{e8e8e8}

\definecolor{C0}{RGB}{31,119,180}
\definecolor{C1}{RGB}{255,127,14}
\definecolor{C2}{RGB}{44,160,44}
\definecolor{C3}{RGB}{214,39,40}
\definecolor{C4}{RGB}{148,103,189}
\definecolor{C5}{RGB}{140,86,75}
\definecolor{C6}{RGB}{227,119,194}
\definecolor{C7}{RGB}{127,127,127}
\definecolor{C8}{RGB}{188,189,34}
\definecolor{C9}{RGB}{23,190,207}

\definecolor{codebg}{rgb}{0.95,0.95,0.95}
\definecolor{codecomment}{rgb}{0.25,0.5,0.35}
\definecolor{codestring}{rgb}{0.6,0.1,0.1}
\definecolor{codekeyword}{rgb}{0.1,0.1,0.7}

\lstdefinestyle{mystyle}{
    backgroundcolor=\color{codebg},
    commentstyle=\color{codecomment}\itshape,
    keywordstyle=\color{codekeyword}\bfseries,
    stringstyle=\color{codestring},
    basicstyle=\ttfamily\small,
    numbers=left,
    numberstyle=\tiny\color{gray},
    stepnumber=1,
    numbersep=8pt,
    showstringspaces=false,
    breaklines=true,
    frame=single,
    rulecolor=\color{gray},
    tabsize=4,
    captionpos=b
}
\renewcommand{\cite}{\citep}

\newcommand{\Uu}{\mathcal{U}}

\newcommand{\EE}{\mathbb{E}}

\DeclareMathOperator{\Pick}{Pick}

\newcommand{\abs}[1]{\left|#1\right|}

\newcommand{\KL}[2]{\mathrm{KL}(#1 \,\|\, #2)}

\newcommand{\correct}{\ensuremath{\#\text{correct}}}
\newcommand{\accepted}{\ensuremath{\#\text{accepted}}}

\newtheorem{theorem}{Theorem}

\title{Parallel Token Prediction \\ for Language Models}

\author{Felix Draxler\textnormal{$^\ast$} \quad Justus Will\thanks{Equal contribution. Corresponding authors: \texttt{\{fdraxler,jcwill,mandt\}@uci.edu}}\,\,\, \quad Farrin Marouf Sofian \\
Department of Computer Science, \\
University of California, Irvine \\
\And
Theofanis Karaletsos \\
Chan-Zuckerberg Initiative \\
\& Pyramidal AI \\
\And
Sameer Singh \\
Department of Computer Science, \\ 
University of California, Irvine \\
\And
Stephan Mandt \\
Computer Science \& AI in Science Institute, \\ 
University of California, Irvine \\
}

\iclrfinalcopy %

\begin{document}

\maketitle

\begin{abstract}
    Autoregressive decoding in language models is inherently slow, generating only one token per forward pass.
    We propose Parallel Token Prediction (PTP), a general-purpose framework for predicting multiple tokens in a single model call.
    PTP moves the source of randomness from post-hoc sampling to random input variables, making future tokens deterministic functions of those inputs and thus jointly predictable in a single forward pass.
    We prove that a single PTP call can represent arbitrary dependencies between tokens.
    PTP is trained by distilling an existing model or through inverse autoregressive training without a teacher.
    Experimentally, PTP achieves a 2.4$\times$ speedup on a diverse-task speculative decoding benchmark.
    We provide code and checkpoints at \mbox{\url{https://github.com/mandt-lab/ptp}}.
\end{abstract}

\section{Introduction}

Autoregressive transformers \citep{vaswani2017attention} are the foundation of today’s large language models (LLMs) \citep{brown2020language}. Their sequential generation process, however, remains a major bottleneck: predicting one token requires one forward pass of the model. This increases inference latency significantly compared to what a single transformer call would achieve.

Many recent efforts aim to bypass this bottleneck by predicting multiple tokens at once. Broadly, they can be categorized into two lines of work: the first, speculative decoding, takes a systems approach, making predictions in a lightweight model that is verified by a large model \citep{leviathan2023fast,chen2023accelerating,sun2023spectr,zhong2025speeding}. The second line of work makes use of predicting several tokens independently of each other.
This significantly reduces the search space for sequences and improves overall model quality
\begin{figure}[h]
    \centering
    \includegraphics[width=.8\linewidth]{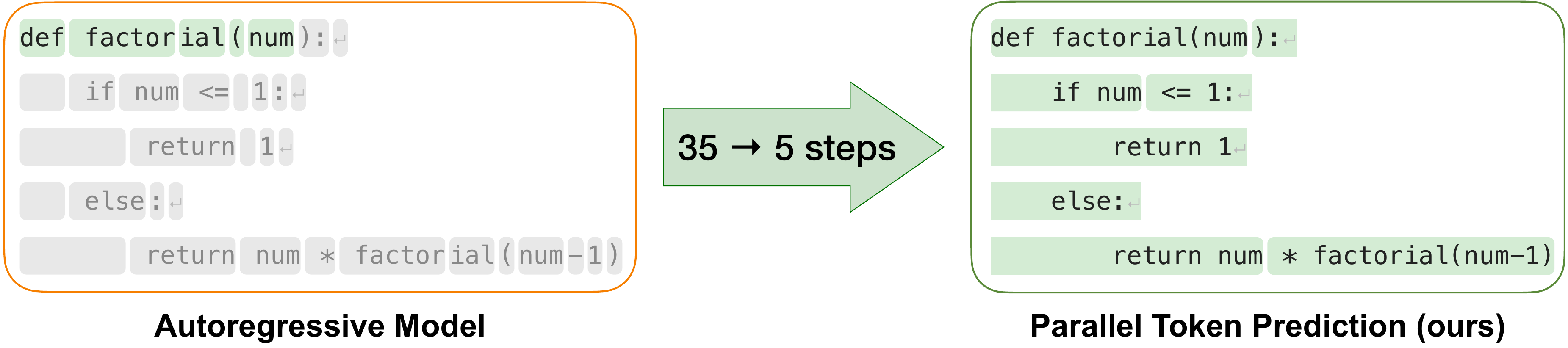}
    \caption{\textbf{Our parallel model generates the same text as its teacher in a fraction of the steps.} By the time our model \textit{(right)} has generated an entire function, its autoregressive teacher \textit{(left)} is still busy with the function header. {\setlength{\fboxsep}{1pt}\colorbox{chipgreen}{Green}} tokens are generated during the time spent on PTP; {\setlength{\fboxsep}{1pt}\colorbox{chipgray}{gray}} tokens show the remaining steps an autoregressive model would still need. Prompt: \texttt{Write a Python function that computes the factorial of a number}.}
    \label{fig:code-sampling}
\end{figure}
\citep{qi2020prophetnet,gloeckle2024better,deepseek-ai2025deepseekv3}.
Similarly, discrete diffusion can iteratively refine generated sequences across several steps \citep{hoogeboom2021argmax,austin2021structured}. However, all of these methods still contain an irreducible sequential component to generate sequences.

Our work takes a step towards filling this gap. We propose a framework that, in theory, can generate arbitrary length sequences in parallel. This is enabled by a small but fundamental architectural change: instead of sampling from the distributions predicted by an autoregressive model in a post-processing step, we feed the involved random variables as an input to the model: the model learns to sample. This enables it to anticipate which tokens will be sampled and to predict them jointly. \Cref{fig:code-sampling} shows how this results in a significant reduction in model calls to produce identical text.
Similar frameworks have been formulated in the normalizing flow literature: Inverse Autoregressive Flows \citep{kingma2016improved} generate samples of many continuous dimensions in parallel and Free-form Flows \citep{draxler2024freeform} distill a fast generator network. We transfer these concepts to sampling discrete sequences from a continuous latent space.

Our contributions are as follows:
\begin{itemize}
    \item We propose \textit{Parallel Token Prediction} (PTP), a modeling approach for discrete data that generates multiple interdependent tokens in one model call~(\cref{sec:parallel-sampling}).
    \item We prove that PTP is as expressive as autoregressive models~(\Cref{thm:one-hot-PTP,thm:C-PTP}).
    \item We propose \emph{Partial Quadratic Decoding}, an efficient error correction scheme that allows parallel verification of long generated sequences during sampling (\cref{sec:error-correction}).
    \item 
    Experimentally, we distill real-world natural language PTP models, achieving a speedup of 2.4$\times$ and 4.2 accepted tokens per speculative decoding step (\cref{sec:experiments}).
\end{itemize}
Together, our framework opens a design space to accurately predict several tokens in parallel, reducing latency in language model output without limiting representational power.

\section{Parallel Token Prediction}

\subsection{Parallel Sampling}
\label{sec:parallel-sampling}

\begin{figure}[h]
    \centering
    \includegraphics[width=0.9\linewidth]{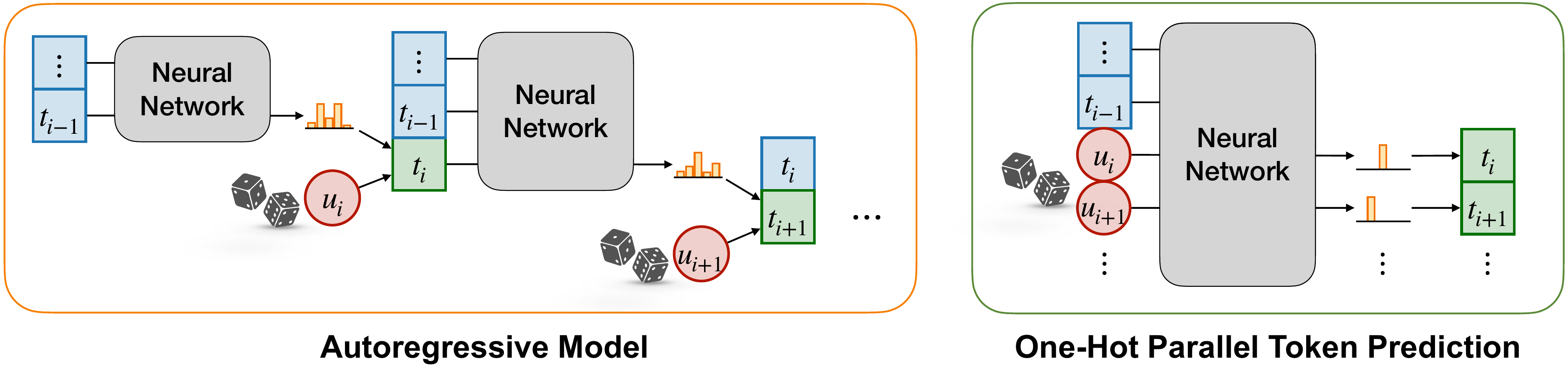}
    \caption{\textbf{Parallel Token Prediction predicts several tokens in one model call.} \textit{(Left)}~An autoregressive model predicts a single $t_i$ per step. It predicts the distribution of each single next token, then the token is chosen with the help of an auxiliary random variable $u_i \sim \Uu[0, 1]$ (see \cref{fig:pick-token-function}). \textit{(Right)}~One-Hot Parallel Token Prediction learns the sampling process by feeding the auxiliaries into the model. This makes predictions deterministic, allowing joint prediction of several tokens. 
    }
    \label{fig:sampling-comparison}
\end{figure}

To construct our \emph{Parallel Token Prediction} framework, let us recap how a classical transformer decoder generates text, see \cref{fig:sampling-comparison} (left). This decoder iteratively predicts the categorical distribution of the next token $t_i \in \{1, \dots, V\}$ based on all previous tokens $t_{<i} = (t_1, \dots, t_{i-1})$,
\begin{equation}
    P_i := P(t_i|t_{<i}).
    \label{eq:autoregressive-model}
\end{equation}
One then samples $t_i \sim P_i$, before moving on to predicting the next token. The requirement to know token $t_i$ to predict the next token $t_{i+1}$ is the reason for the latency of autoregressive inference. For simplicity, we assume this distribution is the final distribution that is used to generate tokens, in that it already reflects temperature scaling \cite{guo2017calibration}, top-k and top-p sampling \cite{holtzman2020curious}, or other approaches trading sample diversity with quality.

To break the autoregressive nature, we need to understand how exactly tokens are sampled from $P_i$: One draws an auxiliary random variable $u_i \sim \Uu[0, 1]$ and looks up the corresponding token from the inverse cumulative distribution function as follows:
\begin{equation}
    t_i = \operatorname{Pick}(u_i, P_i) \equiv \min_{j \in \{1, \dots V\}} \{ j: F_{ij}> u_i \},\quad\text{where } F_{ij} = \sum_{l=1}^{j} P_{il}.
    \label{eq:autoregressive-sampling}
\end{equation}
\begin{wrapfigure}{r}{0.45\textwidth}
    \centering
    \includegraphics[width=\linewidth]{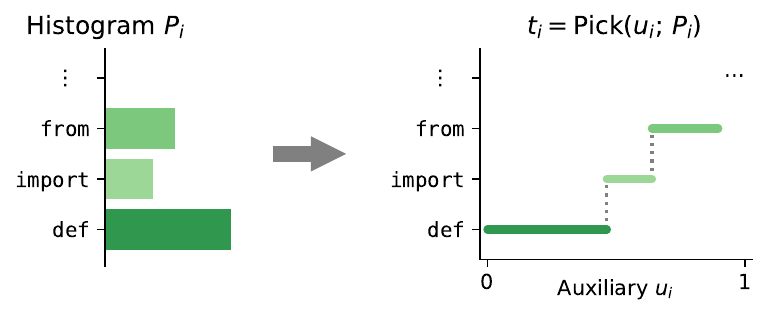}
    \caption{\textbf{Sampling from a discrete distribution.} Given a histogram $P_i$ \textit{(left)}, compute the inverse cumulative distribution function \textit{(right)} and look up the token at a random location $u_i \in \Uu[0, 1]$. PTP learns the function on the right instead of the histogram.}
    \label{fig:pick-token-function}
\end{wrapfigure}
\Cref{fig:pick-token-function} illustrates how this function jumps from token to token as a function of $u_i$.
Here, $j$ iterates possible token choices, $P_{il}$ is the probability to sample $t_i = l$, and $F_{ij}$ is the cumulative distribution to sample a token $t_i \in \{1, \dots j\}$.

Note that while \cref{eq:autoregressive-model} defines a distribution over possible next tokens, \cref{eq:autoregressive-sampling} is a deterministic rule once the auxiliary variable $u_i$ is drawn. Thus, write this rule as an explicit deterministic function:
\begin{equation}
    \label{eq:deterministic-token-given-auxiliary}
    t_i = f_P(t_{<i}; u_i) = \Pick(u_i, P(\,\cdot\,  \mid t_{< i})).
\end{equation}
These auxiliary variables are all we need to perform parallel generation of text: all information about which token $t_i$ we are going to select is available to the model if it has access to $u_i$ as one of its inputs. Since $u_i$ uniquely determines $t_i$ given $t_{<i}$, it carries the same information as $t_i$ itself. The model can therefore use $u_i$ as a stand-in for $t_i$ when predicting $t_{i+1}$, then $u_{i+1}$ as a stand-in for $t_{i+1}$, and so on. Feeding all auxiliary variables $u_i, \dots, u_k$ into the model at once thus allows every future token to be predicted without any sequential dependency (formal proof in \cref{sec:proof-one-hot-PTP}):
\begin{theorem}
    \label{thm:one-hot-PTP}
    Let $P$ denote a probability distribution for next token prediction. Then, the future token $t_k$ can be selected as a deterministic function $f_P$ of previous tokens $t_{<i}$ and auxiliary variables $u_i, \dots, u_{k} \sim \mathcal{U}[0, 1]$:
    \begin{equation}
        t_{k} = f_P(t_{<i}; u_i, \dots u_{k}), \qquad \text{for all }k \geq i.
        \label{eq:one-hot-PTP}
    \end{equation}
\end{theorem}
\Cref{thm:one-hot-PTP} shows a clear path to build a model that can sample many tokens in parallel: instead of learning the distribution $P(t_k \mid t_{<k})$, we propose to directly fit the function $f_P(t_{<i}; u_i, \dots, u_{k})$, which jointly predicts future tokens $t_k$.

\Cref{fig:sampling-comparison} (right) visualizes how this path can be implemented with a standard transformer backbone \cite{vaswani2017attention}: alongside the previous tokens, simply feed random auxiliary variables $u_i, \dots, u_N \sim \Uu[0, 1]$ into the model. Then predict $P(t_k \mid t_{<i}; u_i, \dots, u_k)$ for each future token in one model call. Per \cref{thm:one-hot-PTP}, since $u_k$ uniquely determines $t_k$, this prediction narrows down to a single token --- a one-hot distribution: $P(t_k \mid t_{<i}; u_i, \dots, u_k) = \mathbf{1}(t_k = f_P(t_{<i}; u_i, \dots, u_k))$. In practice, we take the $\mathrm{argmax}$ to extract that predicted token:
\begin{equation}
    \label{eq:o-ptp-sampling}
    t_k = f_P^{\text{O-PTP}}(t_{<i}; u_i, \dots, u_k) = \operatorname{argmax}(P(t_k \mid t_{<i}; u_i, \dots, u_k))
\end{equation}
We refer to this model as a \textbf{One-Hot Parallel Token Prediction Model}~(O-PTP). O-PTPs can be trained to replicate an existing autoregressive model $P$, as we discuss in \cref{sec:distillation}.

O-PTPs converge to one-hot distributions, and therefore do not expose the underlying sampling distributions of the model. While we gain access to the model's confidence, this prevents access to the original conditional probabilities $P(t_k \mid t_{<k})$, which are required for training without a teacher, adjusting temperature \cite{holtzman2020curious}, and uncertainty quantification. To address this, we introduce \textbf{Categorical Parallel Token Prediction}~(C-PTP), which recovers the full conditional distribution of each token. The key idea is to predict each token $t_k$ while conditioning on all \emph{past} auxiliary variables $u_i, \dots, u_{k-1}$, but explicitly excluding its own auxiliary variable $u_k$. By \cref{thm:one-hot-PTP}, these past auxiliaries deterministically encode the sampled history $t_{<k}$. Withholding $u_k$ preserves the uncertainty over $t_k$ rather than collapsing it to a point mass. As a result, conditioning $t_k$ on $(t_{<i}, u_i, \dots, u_{k-1})$ exactly recovers the original autoregressive conditional $P(t_k \mid t_{<k})$ (proof in \cref{sec:proof-C-PTP}):
\begin{theorem}
    \label{thm:C-PTP}
    Let $P$ denote a probability distribution for next token prediction. Then, the distribution of a token $t_{k}$ is fully determined by context tokens $t_{<i}$ and the past auxiliary variables $u_{i}, \dots, u_{k - 1}$:
    \begin{equation}
        \label{eq:C-PTP}
        P(t_{k} \mid t_{<i}, u_i, \dots, u_{k - 1}) = P(t_{k} \mid t_{<k}), \qquad \text{for all }k \geq i.
    \end{equation}
\end{theorem}
\begin{wrapfigure}{r}{0.38\textwidth}
    \centering
    \includegraphics[width=\linewidth]{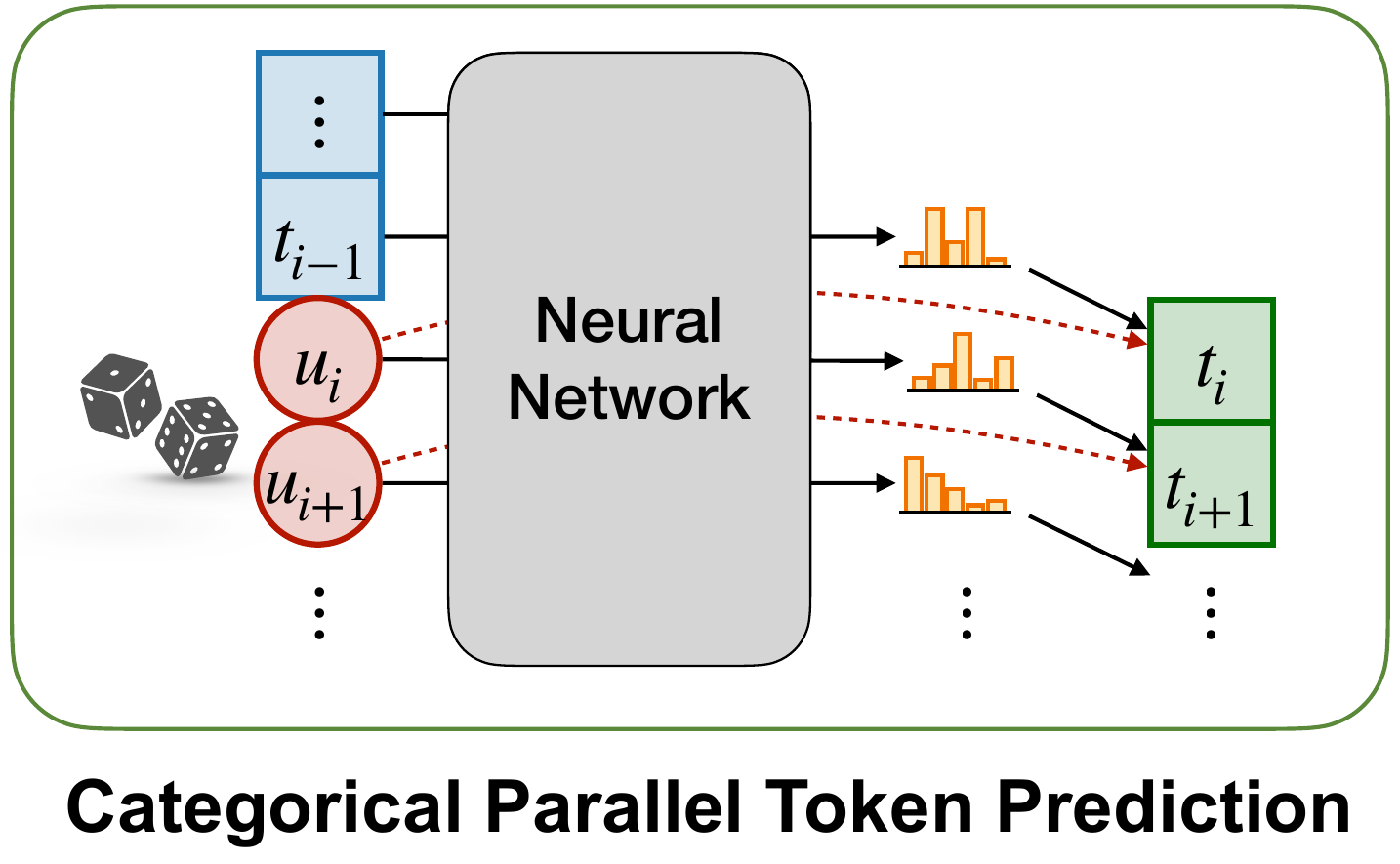}
    \caption{\textbf{Categorical Parallel Token Prediction.} By withholding $u_k$ for the prediction of $t_k$, C-PTP learns the same distribution as an autoregressive model (\cref{thm:C-PTP}). The auxiliary $u_k$ is used to sample token $t_k$.}
    \label{fig:c-ptp}
\end{wrapfigure}
Note how \cref{eq:C-PTP} does not depend on $u_k$. \Cref{fig:c-ptp} shows how \cref{thm:C-PTP} can be used to predict the distribution of the tokens $t_{i}, \dots t_N$ in parallel. Just like for O-PTP, first sample all required auxiliary variables $u_{i}, \dots u_{N} \sim \Uu[0, 1]$, and then predict all $P_k = P(t_{k} \mid t_{<i}, u_i, \dots, u_{k - 1})$ in parallel. Sampling from these distributions is done just as in \cref{eq:deterministic-token-given-auxiliary}:
\begin{align}
    \nonumber
    t_k 
    &= f_{P}^{\text{C-PTP}}(t_{<i}, u_i, \dots, u_{k}) \\ \label{eq:c-ptp-sampling}
    &= \Pick(u_k; P(t_k \mid t_{<i}, u_i, \dots, u_{k-1})).
\end{align}
By using a causal decoder architecture, we can properly mask which token has access to which auxiliaries.

C-PTP can be trained without a teacher by iteratively solving \cref{eq:C-PTP} for the auxiliary $u_k$ corresponding to a given token $t_k$, see \cref{sec:autoregressive-training}. Like O-PTP, it can also be distilled from a teacher.

Both One-Hot and Categorical Parallel Token Prediction allow coordinated token prediction in a single model call. \textbf{By \Cref{thm:one-hot-PTP,thm:C-PTP}, there are no fundamental restrictions as to how many tokens can be jointly modeled} apart from model capacity. In the next section, we propose approaches for training from scratch (only C-PTP) and distilling an existing model (both O-PTP and C-PTP).

\subsection{Training}
\label{sec:training}

Before deriving the training paradigms for Parallel Token Prediction Models, let us quickly recall that autoregressive models are trained by minimizing the cross-entropy between samples from the training data $t \sim P(t)$ and the model $P_\theta$:
\begin{align}
\label{eq:ce}
    \mathcal{L}(\theta) = \EE_{t \sim P(t)}\!\left[- \sum_{i=1}^T \log P_\theta(t_i \mid t_{<i})\right].
\end{align}
Using a causal model such as a transformer \cite{vaswani2017attention}, this loss can be evaluated on an entire sequence of tokens in a single model call \cite{radford2018improving}.
We now present how to distill both One-Hot and Categorical Parallel Token Prediction Models from a trained autoregressive model.
We then show how the categorical variant can be self-distilled from data alone via \cref{eq:ce}.

\subsubsection{Distillation}
\label{sec:distillation}

To build a PTP model $P_\theta$ that speeds up inference of a pretrained autoregressive model $Q_\phi$, we propose to train it using distillation. Given enough model capacity, this results in a fast model that generates identical text as the base model (\Cref{thm:one-hot-PTP,thm:C-PTP}). We defer efficiently correcting errors arising from finite resources to the subsequent \cref{sec:error-correction}.

To train the student for a given training sequence $t = t_{1}, \dots, t_{T}$, we reverse engineer the auxiliary variables $u_1, \dots, u_T$ under which the teacher would have generated the sequence. We then randomly split the sequence into context and prediction sequences, and evaluate a loss that leads the student towards the correct generation. This process is summarized in \cref{alg:ptp-distillation} in \cref{sec:algorithms}.

\textbf{Auxiliary variables.} We train the model based on auxiliary variables that the teacher model would use to generate the training sequence. To this end, we evaluate the teacher distributions of each training token to get the cumulative discrete distributions $F_1, \dots, F_T$ for each token. Inverting \cref{eq:autoregressive-sampling} for $u_k$, we find for every token in the sequence ($k = 1, \dots, T$):
\begin{equation}
    u_{k} \in [F_{k, t_{k}-1}, F_{k, t_{k}}) = \Pick^{-1}(\{t_k\}, P_k).
    \label{eq:compatible-auxiliaries}
\end{equation}
Since $u_{k}$ is continuous, while $t_{k}$ is discrete, we can randomly pick any compatible value. See \cref{sec:beta} for details.

\textbf{Loss evaluation.} Both O-PTP and C-PTP can be trained with standard losses given an input sequence~$t$, with $u_k$ extracted using \cref{eq:compatible-auxiliaries}. We randomly split the sequence at an index~$i$ into context and prediction, and evaluate the following cross-entropy losses, choosing $N$ the index of the last token in the completion part:
\begin{align}
    \label{eq:o-ptp-cross-entropy}
    \text{O-PTP:}&\quad\mathcal{L}(\theta; t, i) = - \sum_{k=i}^{N} \log P_\theta(t_{k} \mid t_{<i}, u_{i}, \dots, u_{k-1}, u_{ k}), \\
    \label{eq:c-ptp-cross-entropy}
    \text{C-PTP:}&\quad\mathcal{L}(\theta; t, i) = - \sum_{k=i}^{N} \log P_\theta(t_{k} \mid t_{<i}, u_{i}, \dots, u_{k-1}).
\intertext{C-PTP can also be trained similar to knowledge distillation \cite{hinton2015distilling}: to this end, we explicitly match the student's $P_{\theta, k}$ to the teacher distribution $Q_{\varphi, k}$. This works for any loss $d(Q, P)$ for the difference between categorical distributions, such as the Kullback–Leibler divergence $d = \KL{Q}{P}$ or its reverse variant $d = \KL{P}{Q}$:}
    &\quad\mathcal{L}(\theta; t, i)
    = \sum_{k=i}^{N} d(Q_\varphi(t_{k} \mid t_{<k}), P_\theta(t_{k} \mid t_{<i}, u_{i \dots k-1})).
    \label{eq:c-ptp-divergence}
\end{align}
At their corresponding minima, the losses in \cref{eq:o-ptp-cross-entropy,eq:c-ptp-cross-entropy,eq:c-ptp-divergence} are minimized by the optimal solutions given in \cref{thm:one-hot-PTP,thm:C-PTP}: a one-hot distribution for O-PTP and the teacher distribution for C-PTP.

\textbf{Training data.} We can train the student using sequences from any data source, even if the teacher assigns different probabilities to them. As long as the teacher assigns non-zero probability and the dataset is sufficiently large, querying the teacher for the corresponding auxiliary variables provides full supervision and allows the student to match the teacher everywhere. We conceptually and empirically compare several training data sources (teacher samples, dataset sequences, student self-samples) in \cref{sec:losses}. We find the lowest-variance option is to sample training sequences from the teacher.

We now show how C-PTP can be trained on training data alone, without a teacher model available.

\subsubsection{Training From Scratch}
\label{sec:autoregressive-training}

Categorical Parallel Token Prediction Models can also be trained directly via \cref{eq:ce}, avoiding the need to have a teacher model as target. For a given training sequence $t_{1}, \dots, t_{T}$, we again split it into the context $t_{<i}$ and the following prediction $t_{\geq i}$. %

Exactly as for distillation, we have to find auxiliary variables that are compatible with every $t_k, k \geq i$. We can do this by selecting, randomly, any
 $   u_{k} \in [F_{k, t_{k}-1}, F_{k, t_{k}})$,
equivalently to \cref{eq:compatible-auxiliaries}, where $F_{k}, t_{k}$ now is the cumulative probability under $P_\theta$ (instead of the teacher model) to choose $t_k$ when predicting that token.
As this probability depends on the previous auxiliary variables $u_{i}, \dots, u_{k-1}$, we select them iteratively.
Specifically, we can alternate between computing the logits of $P_\theta(t_k \mid t_{<i}, u_{i} , \dots, u_{k-1})$, and drawing $u_k$ using \eqref{eq:compatible-auxiliaries}. 

Finally, we can train our model using the cross-entropy loss
\begin{align}
\label{eq:ce-iat}
    \mathcal{L}(\theta) = \EE_{t \sim P(t), i \sim P(i \mid t)}\!\left[- \sum_{k=i}^{N} \log P_\theta(t_{k} \mid t_{<i}, u_{i}, \dots, u_{k-1})\right].
\end{align}
\Cref{alg:ptp-inverse-autoregressive} in \cref{sec:algorithms} summarizes the procedure. A similar approach of iteratively determining latent variables (our auxiliaries) was proposed by Inverse Autoregressive Flows~\cite{kingma2016improved}, although they considered continuous variables in an invertible neural network.

\section{Error Correction}
\label{sec:error-correction}

\Cref{thm:one-hot-PTP,thm:C-PTP} say that O-PTP and C-PTP can in principle predict coherent sequences of arbitrary length in one model call. Practically, finite model capacity limits the length at which a single transformer pass can produce coordinated text. In this section, we propose how to generate identical text to a base model (of arbitrary architecture and size) at minimum latency.

One successful strategy comes in the form of \textit{Speculative Decoding} \cite{leviathan2023fast}. A PTP model proposes several tokens in one model call. A subsequent call to the base model then verifies them. A proposed token $t_k$ is considered correct if it matches $\tilde{t}_k = \mathrm{Pick}(u_k, Q_\varphi(t_k \mid t_{<k}))$, the token the base model selects when the same auxiliary $u_k$ is applied to its own distribution. Additionally, since the base model already evaluates its distribution $Q_\varphi(\cdot \mid t_{<k})$ at every position during verification, one token can be sampled from it for free at the first rejection, resulting in $\accepted{} = \correct{} + 1$ tokens (see \cref{accepted-tokens-metric} for formal definitions). We will find in \cref{sec:speculative-decoding} that PTP models outperform autoregressive models of the same size in terms of wall-clock speedup in vanilla speculative decoding since they only call the student model once.

\label{sec:partial-quadratic-decoding}
Speculative decoding executes proposal and base models in sequence. To reduce this latency, we execute verification and drafting in parallel using \textit{Quadratic Decoding} \cite{samragh2025your}. Since verification has not yet completed when we begin the next drafting step, we do not know how many of the $N_\text{prop}$ previously proposed tokens are correct. Rather than waiting, we prepare for all possible outcomes by constructing a branch for each $m \in \{0, \dots, N_\text{prop}\}$, each assuming exactly $m$ proposed tokens are correct and predicting continuation tokens conditioned on that assumption. All branches are computed concurrently with verification. Once the base model determines the true number $m^*$ of correct tokens, all other branches are discarded and generation continues from the precomputed continuation of branch $m^*$. Parallelization can be achieved with separate hardware or in one merged call by leveraging \emph{Gated Low-Rank Adaption} \cite{samragh2025your} to let different model weights attend to different positions of the input.

\begin{figure}[bt]
    \centering
    \includegraphics[width=0.9\linewidth]{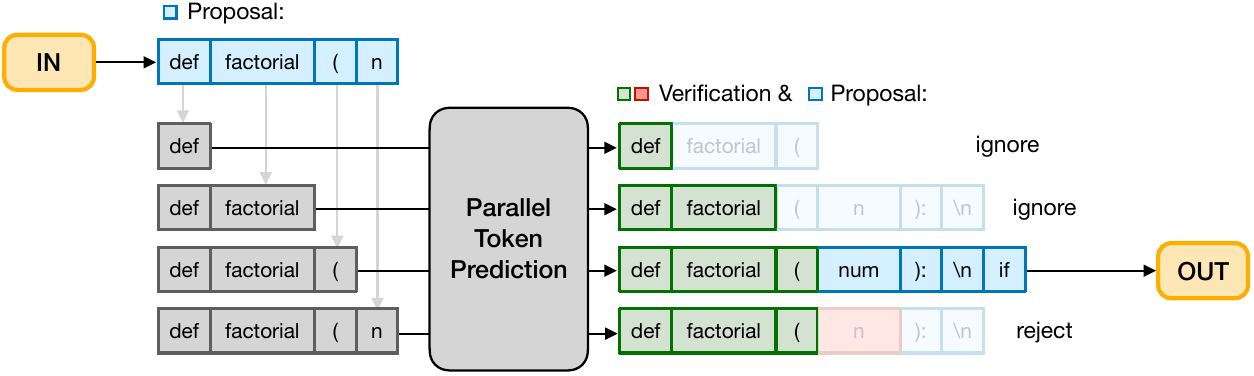}
    \caption{\textbf{Partial Quadratic Decoding performs parallel verification and proposing.} Given a proposal of $N_\text{prop}$ tokens, the model evaluates multiple branches in parallel, each assuming a different number of proposed tokens are correct. {\color{C2}Green} tokens are verified, {\color{C3}red} tokens mark the first mismatch, and {\color{C0}blue} tokens are new PTP proposals. Compute is allocated based on proposal confidence, and the branch with the longest verified prefix is selected.}
    \label{fig:error-correction}
\end{figure}

While quadratic decoding works well for short proposal lengths, its quadratic cost in the number of predicted tokens becomes prohibitively expensive in our case, where we can predict a larger number of coherent tokens. We therefore propose \textbf{Partial Quadratic Decoding}, leveraging confidence estimates from the proposal model to significantly reduce the number of proposed tokens while retaining a high number of accepted tokens per step. \Cref{fig:error-correction} illustrates our proposed decoding scheme, details can be found in \Cref{alg:error-correction-quadratic} and \Cref{app:error-correction}.

The confidence estimates for Partial Quadratic Decoding come directly from the O-PTP model: for each proposed token $t_k$, the model's output probability serves as a confidence score $c_k$, which closely tracks the base model's acceptance rate (see \cref{fig:o-ptp-calibration} in \cref{app:error-correction}).

Given context $t_{<i}$ and proposed tokens $t_k$ for $k \geq i$, with confidences $c_k$ for $k \geq i$, we treat errors as independent and estimate branch probabilities as
\begin{align}
    P(\correct = m \mid t)    &\approx (1 - c_{i+m}) \prod_{k=i}^{i+m-1} c_k \quad \text{for } m < N_\text{prop}.  %
\end{align}
Given a budget of $C$ proposed tokens across all branches, we allocate $B_m$ continuation tokens to branch $m$ by maximizing the expected utility
\begin{equation}
\label{eq:opt-fixed}
    \max_{B \in \{0, \dots, N_\text{prop}\}^{N_\text{prop}+1},\, \sum_m B_m = C} \; \sum_m P(\correct = m \mid t)\, H(B_m),
\end{equation}
where $H(n)$ is a monotonically increasing reward for predicting $n$ tokens. We solve \cref{eq:opt-fixed} greedily, repeatedly assigning the next token to the branch with the highest marginal gain $\Delta_m \coloneqq P(\correct = m \mid t_{\leq k})[H(B_m + 1) - H(B_m)]$. We find $H(n) = \sum_{j=1}^n j\, P(\correct = j)$ works well empirically, with $P(\correct)$ estimated over the training dataset. When additional hardware allows sampling multiple independent proposal sequences in parallel, the acceptance rate can be improved further; see \cref{sec:more_gpu}.

\section{Experiments}
\label{sec:experiments}

We empirically evaluate Parallel Token Prediction (PTP) by its wall-clock speedup compared to an autoregressive model. Since this is hardware-specific, we also report the number of accepted tokens per speculative decoding step (\accepted{} in \cref{sec:error-correction}), as this is a hardware-independent and method-agnostic metric.

We first show that C-PTP can be trained from data alone without access to a teacher model~(\cref{sec:inverse-autoregressive-experiment}). We then distill a 1.1B-parameter model on code generation and demonstrate that O-PTP achieves larger speedups than autoregressive speculative decoding by enabling the draft model to predict multiple tokens per call~(\cref{sec:speculative-decoding}). On the same task, we show that conditioning on auxiliary variables yields substantially more correct tokens than independently predicting multiple tokens~(\cref{sec:independent-failure}). Finally, on a speculative decoding benchmark spanning diverse language tasks, we finetune a 7B-parameter model and show that O-PTP outperforms competitive baselines, achieving a wall-clock 2.4$\times$~speedup~(\cref{sec:vicuna}).

\paragraph{Implementation}

We implement PTP by adding an embedding scheme for the continuous auxiliary tokens $u$ to a standard transformer:
\begin{equation}
    \operatorname{embed}(u) = W \operatorname{binary}(u) + b, \quad\text{where }\operatorname{binary}(u) \in \{0, 1\}^{32}.
\end{equation}
Here, $\operatorname{binary}(u)$ maps the \texttt{float32} number $u$ into the 32 digits of a binary float \cite{witten1987arithmetic}. The linear layer $(W, b)$ maps this binary vector to the embedding space of the autoregressive transformer. 
For distilling a fast student model from a teacher, we train O-PTP with cross-entropy (\cref{eq:o-ptp-cross-entropy}), sample training data from the teacher once and train multiple epochs on it.
We identify these choices through an ablation and practical considerations in \cref{app:ablation}.

\subsection{Inverse Autoregressive Training}
\label{sec:inverse-autoregressive-experiment}

\begin{wraptable}{r}{0.5\textwidth}
    \centering
    \begin{tabular}{cc}
        \toprule
         Model & Perplexity ($\downarrow$) \\
         \midrule
         C-PTP (Ours) & 19.88 \\
         Autoregressive & 19.81 \\
         \bottomrule
    \end{tabular}
    \caption{
    \textbf{Categorical Parallel Token Prediction (C-PTP) can be successfully trained from data alone.} The perplexity of C-PTP and an autoregressive model on taxi pickup location sequences (NYC TLC, \citeyear{nyc_tlc_2016}) are almost identical.
    }
    \label{tab:perp}
\end{wraptable}

We first confirm that C-PTP can be trained as a standalone generative model directly from data. We follow the procedure in \cref{sec:autoregressive-training} and train a model on a dataset that predicts sequences of pick-up locations for taxis in New York City (NYC TLC, \citeyear{nyc_tlc_2016}).

\Cref{tab:perp} shows that the resulting model matches the performance of an autoregressive baseline. Based on the evidence for multi-token prediction \cite{gloeckle2024better,deepseek-ai2025deepseekv3}, we expect the quality of PTP models to outperform autoregressive baselines at larger scales. We leave training larger models from scratch to future work.

\subsection{Limitations of Competing Frameworks}
\label{sec:competition-limitations}

Parallel Token Prediction overcomes important limitations in other frameworks to decrease latency: first, small surrogate models for speculative decoding do not leverage their parallel multi-token potential, which we unlock with our framework in \cref{sec:speculative-decoding}. Second, modeling tokens independently produces incoherent sequences (\cref{sec:independent-failure}). This also limits the representational power of discrete diffusion models \cite{hoogeboom2021argmax,austin2021structured}.
We compare to speculative decoding and independent token prediction on CodeContests \cite{li2023starcoder}, a dataset of coding challenges, by distilling \texttt{TinyLlama-1.1B-Chat-v1.0} \cite{zhang2024tinyllama}.

\subsubsection{Smaller Autoregressive Decoders}
\label{sec:speculative-decoding}

\begin{figure}
    \centering
    \includegraphics[width=.9\linewidth]{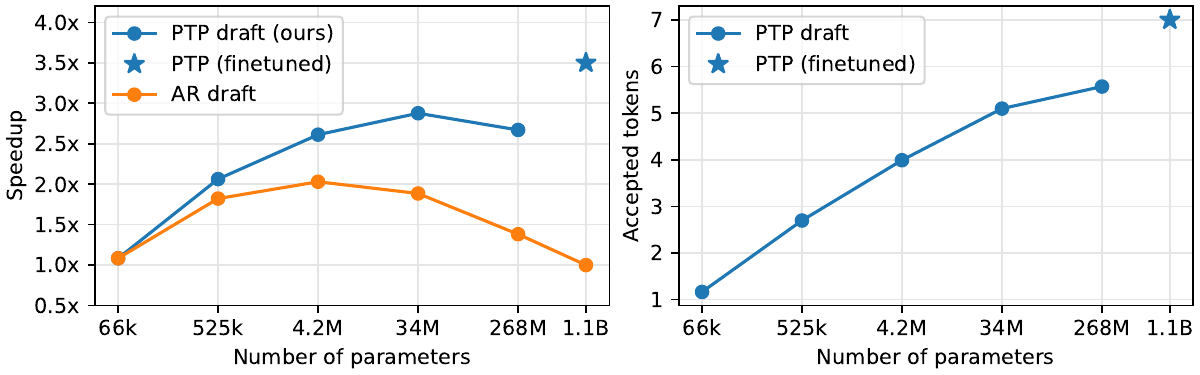}
    \caption{
    \textbf{PTP allows parallelism in draft models for speculative decoding, yielding larger speedups at equal model size.}
    (Left) Wall-clock speedup relative to standard autoregressive decoding (Right) Average number of tokens accepted per step under teacher verification. Curves compare autoregressive draft models to Parallel Token Prediction (PTP) drafts; ($\bullet$) are models trained from scratch for a fixed number of epochs, and ($\star$) is finetuned from the teacher.
    Across model sizes, PTP drafts achieve higher speedups by generating more than one correct token per step, whereas autoregressive drafts remain sequential. 
    }
    \label{fig:speculative-decoding}
\end{figure}

Vanilla speculative decoding distills small autoregressive draft models to imitate the teacher with fewer parameters \cite{leviathan2023fast}. Through their reduced size, these models require less compute to predict each token, and the teacher is used to verify them in parallel as in \cref{sec:error-correction}.

\Cref{fig:speculative-decoding} shows that PTP draft models outperform AR draft models because they need fewer student calls for the same number of tokens.
This shifts the optimal model size towards larger models. In fact, we find that the best performance is achieved by finetuning the teacher model directly. The latter may be an artifact of the smaller models being trained from scratch instead of being finetuned.

\subsubsection{Independent Prediction}
\label{sec:independent-failure}

\begin{wraptable}{r}{0.45\textwidth}
   \centering
   \begin{tabular}{cc}
       \toprule
       Parallelization technique & \accepted{} ($\uparrow$) \\
       \midrule
       O-PTP (ours) & \textbf{7.0 $\pm$ 0.1} \\
       Independent prediction & 6.2 $\pm$ 0.1 \\
       \bottomrule
   \end{tabular}
   \caption{\textbf{Speculative decoding with O-PTP accepts more tokens per step than independently predicting tokens.} We distill a draft model once with meaningful auxiliaries that determine token dependence and once with a generic \texttt{[MASK]} token. Error is difference between two runs.}
   \label{tab:codecontest}
\end{wraptable}
Many approaches to predicting multiple tokens in parallel, including multi-token prediction \cite{qi2020prophetnet,gloeckle2024better} and discrete diffusion models \cite{hoogeboom2021argmax,austin2021structured}, assume that future tokens are conditionally independent. As a result, later tokens are sampled from marginal distributions that average over incompatible earlier choices, making semantic and syntactic inconsistencies unavoidable even with infinite model capacity. In code generation, this manifests as spurious token combinations such as \texttt{def numpy} and \texttt{import find}. See \cref{app:independent-failure-formal} for a formal derivation. 

\begin{figure}[bt]
    \centering
    \includegraphics[width=\linewidth]{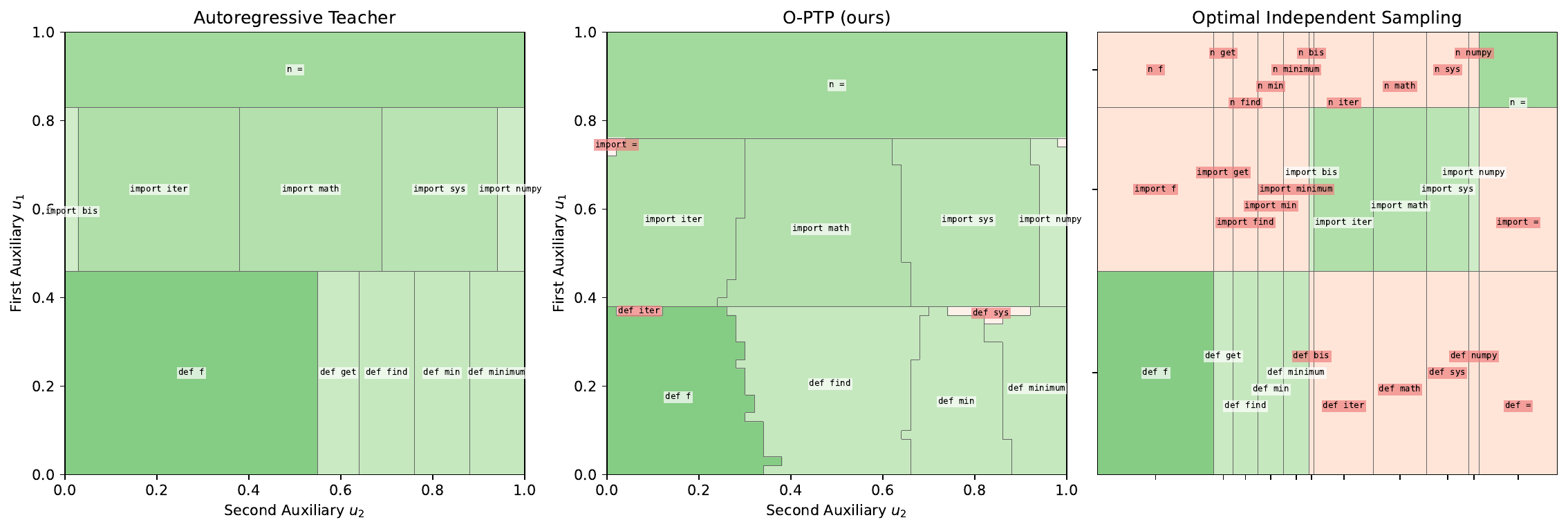}
    \caption{
    \textbf{Parallel Token Prediction generates meaningful pairs of tokens.}
    Each panel plots the first and second auxiliary variables $(u_1,u_2)\in[0,1]^2$ on the axes; regions in this unit square correspond to the resulting two-token outputs (green: compatible, red: incompatible).
    \textit{(Left)} In a coding problem, autoregressive sampling generates meaningful token pairs, such as importing a package or defining a method.
    \textit{(Center)} Our O-PTP coding model produces sensible token pairs, but in a single model call; only rarely ($<1\%$) does it yield spurious combinations like \texttt{def sys}.
    \textit{(Right)} A model that independently predicts future tokens is bound to fail: in about $60\%$ of cases, it combines incompatible tokens because the second token is not informed about the first.
    }
    \label{fig:independent-failure}
\end{figure}

To isolate this effect, we compare models trained with informative auxiliary variables $u_i$ (our O-PTP) to models that have multiple independent heads for predicting multiple tokens~\cite{qi2020prophetnet,gloeckle2024better}.
As shown in \Cref{fig:independent-failure}, O-PTP consistently produces meaningful token combinations by coordinating predictions through the auxiliary variables, while independent prediction frequently produces incompatible pairs. Quantitatively, \cref{tab:codecontest} shows that auxiliary variables substantially increase the number of tokens when used in sequential speculative decoding. \Cref{fig:code-sampling} shows a qualitative sample of our model's predictions.

\subsection{General-purpose Text Generation}
\label{sec:vicuna}

In the following experiment, we confirm that we can finetune a large pretrained model using PTP, resulting in a model that predicts several tokens per forward pass.

We distill a One-Hot Parallel Token Prediction model (O-PTP) student from an  autoregressive teacher, following the procedure in \cref{sec:distillation}. We use Vicuna-7B \cite{vicuna2023} as the teacher and finetune the student on ShareGPT conversational data \cite{chen2024sharegpt4v}. Instead of finetuning the full model, we train a gated LoRA \cite{samragh2025your} adapter, allowing us to make only the minimal changes that are necessary to correctly parse the auxiliary variables.

To reflect a practical deployment scenario for a chat-oriented large language model, we evaluate the performance on SpecBench \cite{xia2024unlocking}, containing a diverse set of tasks: multi-turn conversation (MTC), translation (TL), summarization (SUM), question answering (QA), mathematical reasoning (Math), and retrieval-augmented generation (RAG). Performance is measured using the wall-clock speedup on a single NVIDIA RTX A6000. We also report the hardware-agnostic average number of accepted tokens, as this is highly correlated to the maximum speedup that can be obtained with efficient decoding schemes.

As shown in \cref{tab:spec-bench}, O-PTP consistently predicts more tokens identical to its teacher than existing parallelization baselines across most tasks, resulting in a larger speedup. These results demonstrate that PTP is effective when scaled to large models and heterogeneous real-world datasets. See \Cref{app:more-tables} for detailed results, and \Cref{app:self-verification} for inference without error correction.

\begin{table}[ht]
    \centering
    \resizebox{\linewidth}{!}{
    \begin{tabular}{lccccccl|l}
        \toprule
        Parallelization & MTC & TL & SUM & QA & Math & RAG & Task Average & \#accepted ($\uparrow$) \\
        \midrule
        O-PTP (ours) & 2.77 & \textbf{2.12} & 2.47 & 1.91 & \textbf{2.79} & \textbf{2.32} & \textbf{2.40 $\pm$ 0.007} & \textbf{4.21 $\pm$ 0.011}$^\text{\textdagger}$\\
        SAMD \cite{hu2025sam} & \textbf{2.79} & 1.87 & 2.47 & 2.05 & 2.60 & 2.22 & 2.34 $\pm$ 0.012 & 3.95 $\pm$ 0.011 \\
        Eagle-2 \cite{li2024eagle} & 2.76 & 1.88 & 2.19 & 2.08 & 2.75 & 2.06 & 2.29 $\pm$ 0.010 & 3.93 $\pm$ 0.008 \\
        Hydra \cite{ankner2024hydra} & 2.59 & 1.98 & 1.91 & \textbf{2.16} & 2.56 & 1.86 & 2.18 $\pm$ 0.007 & 3.40 $\pm$ 0.005 \\
        Eagle \cite{li2024eagle1} & 2.50  & 1.80 & 2.10 & 1.93 & 2.49 & 1.94 & 2.13 $\pm$ 0.010 & 3.24 $\pm$ 0.006 \\
        Recycling \cite{luo2025turning} & 2.08 & 1.77 & 1.93 & 1.83 & 2.21 & 1.73 & 1.93 $\pm$ 0.007 & 2.38 $\pm$ 0.003 \\
        Medusa \cite{cai2024medusa} & 2.08 & 1.69 & 1.57 & 1.76 & 2.07 & 1.55 & 1.79 $\pm$ 0.004 & 1.93 $\pm$ 0.003 \\
        PLD \cite{saxena2023prompt} & 1.65 & 1.08 & \textbf{2.50} & 1.20 & 1.62 & 1.75 & 1.63 $\pm$ 0.004 & 2.70 $\pm$ 0.006 \\
        SpS \cite{chen2023accelerating} & 1.57 & 1.12 & 1.54 & 1.39 & 1.40 & 1.55 & 1.43 $\pm$ 0.006 & 1.73 $\pm$ 0.005 \\
        Lookahead \cite{fu2024break} & 1.46 & 1.13 & 1.26 & 1.24 & 1.55 & 1.17 & 1.30 $\pm$ 0.005 & 1.57 $\pm$ 0.003 \\
        None & 1.00 & 1.00 & 1.00 & 1.00 & 1.00 & 1.00 & 1.00 & 1.00 \\
        \bottomrule
    \end{tabular}
    }
    \caption{\textbf{Wall-clock speedups under teacher verification ($\uparrow$).} We compare our One-Hot Parallel Token Prediction (O-PTP) distilled from Vicuna-7B~\cite{vicuna2023} to concurrent parallelization techniques on the diverse tasks of SpecBench~\cite{xia2024unlocking}, see main text for abbreviations. Errors indicate the standard error over three runs. Temperature is set to $0.7$. Across most tasks, O-PTP achieves the largest speed-up.
    We also report \accepted{} ($\uparrow$), the number of accepted tokens per base model call. Here, the errors are standard deviations over all steps across three runs.
    $^\text{\textdagger}$Parallel verification trades number of accepted tokens for total speedup so we report sequential verification here. For more detailed results see \Cref{app:more-tables}.}
    \label{tab:spec-bench}
\end{table}

\section{Related Work}

Speeding up the generation of autoregressive models and discrete sequence models in particular has been the focus of a broad body of work, see \cite{khoshnoodi2024comprehensive} for an overview.

Our framework combines two ideas from the Normalizing Flow literature and imports them to modeling discrete data: Inverse Autoregressive Flows (IAF) are trained with fast prediction in mind \cite{kingma2016improved} by iteratively identifying latent variables (our auxiliary variables) that generate a particular continuous one-dimensional value, and Free-Form Flows (FFF) train a single-step generating function as the inverse of an encoder \cite{draxler2024freeform}.

In the LLM literature, speeding up generation has been approached from various angles. \textbf{Speculative decoding} takes a system perspective, using a small draft model to propose multiple tokens and a large target model to verify them \cite{leviathan2023fast,chen2023accelerating}. Variants verify entire sequences 
\cite{sun2023spectr} or use a smaller verifier network \citet{zhong2025speeding} to improve quality and speed. \textbf{Latent variable methods} first sample latent codes from the prompt so that the distribution of subsequent tokens factorizes given latent codes \cite{gu2018nonautoregressive,ma2019flowseq}.
\textbf{Discrete Diffusion Models} leave autoregressive sampling behind by iteratively refining the text starting from a noisy or masked variant \cite{hoogeboom2021argmax,austin2021structured}. \textbf{Multi-head output models} predict several next tokens independent of each other \cite{qi2020prophetnet,gloeckle2024better,deepseek-ai2025deepseekv3}, narrowing down on the possible set of next tokens. Both discrete diffusion and multi-head models assume independence of tokens, which is fundamentally limited in modeling capacity, compare \cref{sec:independent-failure} and \citet{feng2025theoretical}. Recent work on discrete diffusion models leverages copula models to capture dependencies in an additional model \cite{liu2025discrete}. We model dependencies between several tokens without an additional model at inference time.

In contrast to the above, our framework is universal in the sense that it can approximate arbitrary dependence between several tokens in a single model call. Our new method is complementary to existing approaches, and we leave exploring these combinations open for future research.

\section{Conclusion}

In this paper, we introduce \textit{Parallel Token Prediction}, a framework that permits consistent generation of several tokens in a single autoregressive model call. It eliminates the independence assumptions that limited prior approaches, allowing one to model multiple tokens with arbitrary dependency between them. Empirically, we show that existing models can be distilled into efficient parallel samplers. With error correction, these models produce identical output as a base model while significantly increasing how many tokens are obtained per model call.

This speedup makes language models more practical for real-time applications. Future work includes extending our framework to large scale models, multimodal generation, combining it with complementary acceleration strategies, and exploring theoretical limits on parallelization.

Overall, our results suggest that the sequential bottleneck in autoregressive transformers is not inherent, and that universal, efficient parallel generation is within reach.
We think that our experiments in \cref{sec:experiments} only scratch the surface of the possibilities enabled by our \cref{thm:one-hot-PTP,thm:C-PTP}. We envision future work to train large models from scratch that think in long sequences, and speculate that planning over longer spans may improve downstream performance.

\paragraph{Limitations} The experiments in the present paper focus on distilling existing models, leaving open the question of whether model quality will improve when training from scratch (as is observed for multi-token prediction \cite{gloeckle2024better}).
The inference speedup only manifests given heavily parallelized hardware, where increasing the number of predicted tokens only marginally increases computational cost.

\section*{Acknowledgments}

Justus Will acknowledges support from the HPI Research Center in Machine Learning and Data Science at UC Irvine.
Sameer Singh acknowledges funding from the National Science Foundation (NSF) through an NSF CAREER award IIS-2046873. Stephan Mandt acknowledges funding from the National Science Foundation (NSF) through an NSF CAREER Award IIS-2047418, IIS-2007719, the NSF LEAP Center, and the Hasso Plattner Research Center at UCI, and the Chan Zuckerberg Initiative.

\section*{Ethics Statement}
Our work focuses on reducing the inference time of Large Language Models, enabling more computations per unit time and supporting large-scale or real-time applications. While this can improve responsiveness and resource efficiency, it may also increase the potential for misuse, such as generating misinformation or automated spam at higher volumes. Faster inference does not mitigate underlying model biases, so responsible deployment, monitoring, and safeguards are critical to balance performance gains with societal risks.

\section*{Reproducibility Statement}

We include proofs for all theoretical results introduced in the main text in \cref{sec:proofs}. We include further experimental and implementation details (including model architectures and other hyperparameter choices) in \cref{app:ablation} and \cref{sec:experimental-details}. Our code is available at \mbox{\url{https://github.com/mandt-lab/ptp}}.

\section*{Usage of Large Language Models (LLMs)}

We used Large Language Models to improve the writing of this paper, by pasting text and asking for feedback and spell checking; we also used them for code editing, in particular bug resolving. We also used LLMs for discovering related work. All ideas presented in this paper are our own.

\bibliography{iclr2026_conference}
\bibliographystyle{iclr2026_conference}

\clearpage
\appendix

\section{Proofs}
\label{sec:proofs}

\subsection{Proof of \texorpdfstring{\cref{thm:one-hot-PTP}}{Theorem \ref{thm:one-hot-PTP}}}
\label{sec:proof-one-hot-PTP}

\begin{proof}
    We prove by induction over $k, k\geq i$.

    For $k = i$, the function is given by \cref{eq:deterministic-token-given-auxiliary}.

    For $k \mapsto k + 1$, assume the statement holds for $i, \dots, k$. This gives us access to all $t_1, \dots, t_{i-1}$ (from the context), and all $t_i, \dots, t_k$ using $u_i, \dots, u_k$ (by induction assumption).

    Then, we can compute $P_{k+1} = P(t_{k+1} \mid t_{<i}, t_i, \dots, t_k)$, and pick the next token deterministically via \cref{eq:autoregressive-sampling}:
    \begin{equation}
        t_{k+1} = \Pick(u_{k+1}; P_{k+1}) = f_P(t_1, \dots, t_{i-1}; u_i, \dots u_k, u_{k+1}).
    \end{equation}
\end{proof}

\subsection{Proof of \texorpdfstring{\cref{thm:C-PTP}}{Theorem \ref{thm:C-PTP}}}
\label{sec:proof-C-PTP}

\begin{proof}
    We prove by induction over $k, k\geq i$.

    For $k=i$, there is nothing to show, since there are no auxiliaries involved in the statement.

    For $k \mapsto k+1$, assume the statement holds for $k$. This gives us access to the distribution $P_{k} = P(t_k \mid t_{<k}) = P(t_k \mid t_{<i}, u_i, \dots u_{k-1})$ of the token $t_{k}$. Since token $t_{k}$ is uniquely determined from $P_{k}$ and $u_{k}$ via \cref{eq:deterministic-token-given-auxiliary}, any distribution conditioning on $P_{k}, t_{k}$ can instead condition on $P_{k}, u_{k}$ via the law of total probability, so that $P(t_{k+1} \mid t_{<i}, u_i, \dots u_{k}) = P(t_{k+1} \mid t_{<k}, t_k)$.
\end{proof}

\section{Formal limitations of independent prediction}
\label{app:independent-failure-formal}

Without the auxiliary variables, the distribution of all tokens $k > i + 1$ is not informed about the choice we make for tokens $t_{i+1}, \dots t_{k-1}$. Instead, they are marginalized out, which explains the spurious code snippets:
\begin{align}
    \text{Independent: }& P(t_{k} \mid t_{<i}) = \sum_{t_{i}, \dots t_{k-1}} P(t_{k} \mid t_{<k}) P(t_{i}, \dots, t_{k-1} \mid t_{<i}) \neq P(t_{k} \mid t_{<k}). \\
    \text{Dependent (ours): }& P(t_{k} \mid t_{<i}, u_i, \dots u_{k-1}) \overset{\text{thm.~\ref{thm:C-PTP}}}{=} P(t_{k} \mid t_{<k}).
\end{align}
This limits how many tokens can be sampled in parallel without auxiliary variables, even for infinite model capacity. Our \cref{thm:one-hot-PTP,thm:C-PTP} allow for coordinating tokens, making model capacity the only restriction.

\section{Average number of correct tokens}
\label{accepted-tokens-metric}

To quantify the effectiveness of parallel token generation, we measure the number of correctly predicted tokens. Given a student output sequence $t_1^{(\mathrm{S})}, t_2^{(\mathrm{S})}, \dots, t_T^{(\mathrm{S})}$ and a teacher output sequence $t_1^{(\mathrm{T})}, t_2^{(\mathrm{T})}, \dots, t_T^{(\mathrm{T})}$ of the same length, this metric is defined as
\begin{align}
    \label{eq:correct-count}
    \correct &= \max \left\{ k \;\middle|\; t_j^{(\mathrm{S})} = t_j^{(\mathrm{T})} \;\; \forall\, j \le k \right\}.
\end{align}
That is, \correct{} equals the length of the longest prefix of the student's output that exactly matches the teacher's output.
We further define \accepted{} as the number of tokens accepted per base model call. Often, for example with parallel verification, we have $\correct = \accepted$. Note, however, that with sequential error correction we obtain
\begin{align}
    \accepted &= \min\{\correct + 1, T\}.
\end{align}
Intuitively, this measures how many tokens the student can generate in parallel before the first disagreement, and directly corresponds to the number of tokens that can be safely accepted without error correction.

\section{Detailed Comparison on SpecBench} \label{app:more-tables}

We report more extensive results from our experiment in \Cref{sec:experiments}. See \Cref{tab:spec-bench-correct} for average number of accepted tokens and \Cref{tab:spec-bench-speed} for wall-clock speed-ups.

\begin{table}[ht]
    \centering
    \resizebox{\linewidth}{!}{
    \begin{tabular}{lllllllll}
        \toprule
        Parallelization & MTC & TL & SUM & QA & Math & RAG & Task-Average \\
        \midrule
        O-PTP (ours) & 2.77 $\pm$ 0.021 & \textbf{2.12 $\pm$ 0.027} & 2.47 $\pm$ 0.019 & 1.91 $\pm$ 0.019 & \textbf{2.79 $\pm$ 0.012} & \textbf{2.32 $\pm$ 0.021} & \textbf{2.40 $\pm$ 0.007} \\
        SAMD \cite{hu2025sam} & \textbf{2.79 $\pm$ 0.026} & 1.87 $\pm$ 0.016 & 2.47 $\pm$ 0.043 & 2.05 $\pm$ 0.007 & 2.60 $\pm$ 0.018 & 2.22 $\pm$ 0.056 & 2.34 $\pm$ 0.012 \\
        Eagle-2 \cite{li2024eagle} & 2.76 $\pm$ 0.049 & 1.88 $\pm$ 0.008 & 2.19 $\pm$ 0.014 & 2.08 $\pm$ 0.015 & 2.75 $\pm$ 0.036 & 2.06 $\pm$ 0.024 & 2.29 $\pm$ 0.010 \\
        Hydra \cite{ankner2024hydra} & 2.59 $\pm$ 0.022 & 1.98 $\pm$ 0.006 & 1.91 $\pm$ 0.010 & \textbf{2.16 $\pm$ 0.011} & 2.56 $\pm$ 0.031 & 1.86 $\pm$ 0.030 & 2.18 $\pm$ 0.007 \\
        Eagle \cite{li2024eagle1} & 2.50 $\pm$ 0.057 & 1.80 $\pm$ 0.009 & 2.10 $\pm$ 0.005 & 1.93 $\pm$ 0.028 & 2.49 $\pm$ 0.013 & 1.94 $\pm$ 0.017 & 2.13 $\pm$ 0.010 \\
        Recycling \cite{luo2025turning} & 2.08 $\pm$ 0.045 & 1.77 $\pm$ 0.008 & 1.93 $\pm$ 0.004 & 1.83 $\pm$ 0.007 & 2.21 $\pm$ 0.010 & 1.73 $\pm$ 0.013 & 1.93 $\pm$ 0.007 \\
        Medusa \cite{cai2024medusa} & 2.08 $\pm$ 0.011 & 1.69 $\pm$ 0.008 & 1.57 $\pm$ 0.002 & 1.76 $\pm$ 0.019 & 2.07 $\pm$ 0.008 & 1.55 $\pm$ 0.015 & 1.79 $\pm$ 0.004 \\
        PLD \cite{saxena2023prompt} & 1.65 $\pm$ 0.024 & 1.08 $\pm$ 0.006 & \textbf{2.50 $\pm$ 0.004} & 1.20 $\pm$ 0.011 & 1.62 $\pm$ 0.005 & 1.75 $\pm$ 0.006 & 1.63 $\pm$ 0.004 \\
        SpS \cite{chen2023accelerating} & 1.57 $\pm$ 0.018 & 1.12 $\pm$ 0.007 & 1.54 $\pm$ 0.006 & 1.39 $\pm$ 0.008 & 1.40 $\pm$ 0.021 & 1.55 $\pm$ 0.023 & 1.43 $\pm$ 0.006 \\
        Lookahead \cite{fu2024break} & 1.46 $\pm$ 0.021 & 1.13 $\pm$ 0.024 & 1.26 $\pm$ 0.005 & 1.24 $\pm$ 0.010 & 1.55 $\pm$ 0.003 & 1.17 $\pm$ 0.007 & 1.30 $\pm$ 0.005 \\
        None & 1.00 & 1.00 & 1.00 & 1.00 & 1.00 & 1.00 & 1.00 \\
        \bottomrule
    \end{tabular}
    }
    \caption{\textbf{Wall-clock speedups under teacher verification ($\uparrow$).} Errors indicate the standard error over three runs. Temperature is set to $0.7$. Across most tasks, O-PTP achieves the largest speed-up.}
    \label{tab:spec-bench-speed}
\end{table}
\begin{table}[ht]
    \centering
    \resizebox{\linewidth}{!}{
    \begin{tabular}{lllllllll}
        \toprule
        Parallelization & MTC & TL & SUM & QA & Math & RAG & Task-Average \\
        \midrule
        O-PTP, sequential (ours) & \textbf{4.52 $\pm$ 0.019} & \textbf{3.73 $\pm$ 0.041} & \textbf{4.53 $\pm$ 0.027} & 3.42 $\pm$ 0.022 & \textbf{4.62 $\pm$ 0.028} & \textbf{4.29 $\pm$ 0.037} & \textbf{4.21 $\pm$ 0.011} \\
        O-PTP, parallel (ours) & 3.00 $\pm$ 0.015 & 2.59 $\pm$ 0.037 & 3.12 $\pm$ 0.024 & 2.23 $\pm$ 0.017 & 3.44 $\pm$ 0.027 & 2.93 $\pm$ 0.032 & 2.89 $\pm$ 0.009 \\
        \midrule
        SAMD \cite{hu2025sam} & \textbf{4.52 $\pm$ 0.023} & 3.16 $\pm$ 0.032 & 4.08 $\pm$ 0.036 & \textbf{3.44 $\pm$ 0.028} & 4.37 $\pm$ 0.024 & 3.86 $\pm$ 0.045 & 3.95 $\pm$ 0.011 \\
        Eagle-2 \cite{li2024eagle} & 4.44 $\pm$ 0.011 & 3.20 $\pm$ 0.031 & 3.79 $\pm$ 0.017 & 3.40 $\pm$ 0.019 & 4.62 $\pm$ 0.021 & 3.89 $\pm$ 0.024 & 3.93 $\pm$ 0.008 \\
        Hydra \cite{ankner2024hydra} & 3.90 $\pm$ 0.006 & 2.88 $\pm$ 0.023 & 2.95 $\pm$ 0.010 & 3.21 $\pm$ 0.013 & 3.90 $\pm$ 0.009 & 3.37 $\pm$ 0.014 & 3.40 $\pm$ 0.005 \\
        Eagle \cite{li2024eagle1} & 3.63 $\pm$ 0.008 & 2.72 $\pm$ 0.024 & 3.17 $\pm$ 0.013 & 2.88 $\pm$ 0.015 & 3.76 $\pm$ 0.014 & 3.14 $\pm$ 0.018 & 3.24 $\pm$ 0.006 \\
        Recycling \cite{luo2025turning} & 2.74 $\pm$ 0.007 & 2.46 $\pm$ 0.029 & 2.68 $\pm$ 0.013 & 2.60 $\pm$ 0.014 & 3.06 $\pm$ 0.014 & 2.60 $\pm$ 0.017 & 2.70 $\pm$ 0.006 \\
        Medusa \cite{cai2024medusa} & 2.70 $\pm$ 0.005 & 2.18 $\pm$ 0.016 & 2.12 $\pm$ 0.007 & 2.24 $\pm$ 0.009 & 2.67 $\pm$ 0.008 & 2.26 $\pm$ 0.010 & 2.38 $\pm$ 0.003 \\
        SpS \cite{chen2023accelerating} & 2.10 $\pm$ 0.006 & 1.40 $\pm$ 0.009 & 2.15 $\pm$ 0.011 & 1.79 $\pm$ 0.008 & 1.89 $\pm$ 0.008 & 2.17 $\pm$ 0.014 & 1.93 $\pm$ 0.003 \\
        PLD \cite{saxena2023prompt} & 1.67 $\pm$ 0.007 & 1.10 $\pm$ 0.007 & 2.73 $\pm$ 0.024 & 1.37 $\pm$ 0.010 & 1.80 $\pm$ 0.012 & 1.71 $\pm$ 0.016 & 1.73 $\pm$ 0.005 \\
        Lookahead \cite{fu2024break} & 1.69 $\pm$ 0.004 & 1.24 $\pm$ 0.010 & 1.54 $\pm$ 0.006 & 1.56 $\pm$ 0.007 & 1.92 $\pm$ 0.008 & 1.42 $\pm$ 0.008 & 1.57 $\pm$ 0.003\\
        None & 1.00 & 1.00 & 1.00 & 1.00 & 1.00 & 1.00 & 1.00 \\
        \bottomrule
    \end{tabular}
    }
    \caption{\textbf{Average number of accepted tokens under teacher verification ($\uparrow$)}. 
    For each method and task, we list the average length of the longest prefix matching the teacher’s output, which directly determines decoding speedup. Errors indicate the standard error over total number of model calls. Temperature is set to $0.7$. Across nearly all tasks, O-PTP achieves the highest number of accepted tokens. Parallel verification trades number of accepted tokens for total speedup, so we report both for our model.}
    \label{tab:spec-bench-correct}
\end{table}

\section{Additional Ablation Results}
\label{app:ablation}

Based on a dataset (NYC TLC, \citeyear{nyc_tlc_2016}) that contains latitudes and longitudes for pick-up locations for all taxi rides in 2016, we divide the city into $25$ neighborhoods via $k$-Means clustering to obtain a discrete-valued time-series that we can split into overlapping chunks of length $N$. This is a common benchmark dataset in the literature of marked temporal point processes \cite{xue2024easytpp}, and autoregressive transformers are a common architecture \cite{draxler2025transformers}.

We now provide some of the specific choices we made when implementing the general framework of Parallel Token Prediction. Specifically, we discuss the empirical difference between O-PTP and C-PTP and which specific loss to choose. We will specify our model architecture and how to embed both tokens and auxiliary variables in the same embedding space, and lastly compare the proposal distributions our training sequences can be sampled from.

We test our framework by training a model that predicts pick-up locations for taxis in New York City. Based on a dataset (NYC TLC, \citeyear{nyc_tlc_2016}) that contains latitudes and longitudes for pick-up locations for all taxi rides in 2016, we divide the city into $25$ neighborhoods via $k$-Means clustering to obtain a discrete-valued time-series that we can split into overlapping chunks of length $N$. This is a common benchmark dataset in the literature of marked temporal point processes \cite{xue2024easytpp}.

As a teacher model, we pretrain a $29$M-parameter autoregressive causal transformer based on the architecture of GPT-2 \cite{radford2019language}, using the cross-entropy loss in \cref{eq:ce}. For our PLM we choose the same GPT-style transformer architecture as the teacher. This allows us to use the teacher's parameters as a warm-start.
We evaluate all our parallel models in terms of the average number of leading tokens predicted by our student model that are identical to the teacher. In the end, this is the quantity that limits the maximum latency reduction that can be achieved, see \cref{sec:error-correction}.

\subsection{Auxiliary Variable Embeddings}

In our experiments we use transformers that embed tokens into a higher-dimensional embedding space via a learned embedding before adding a positional embedding. This doesn't work out-of-the-box for our auxiliary variables since they are one-dimensional continuous variables. Thus we learn a separate embedding. We combine two components, for each of which we test several variants:
(1)~A learned affine linear transform [\textbf{lin}] or a fully connected neural network [\textbf{NN}]. (2)~Feed either the scalar $u$ [\textbf{fl}], an $n$-dimensional threshold-embedding $e_j = 1\{u \leq j/n\}$ [\textbf{th}], or an $n$-dimensional embedding $e_j = 1\{u 2^{j-1} \,\mathrm{mod}\, 1 \leq 0.5\}$ [\textbf{ar}] inspired by arithmetic coding \cite{witten1987arithmetic}.

Empirically, all methods work reasonably well, but a structured embedding leads to faster and more stable training convergence. This is similar to the transformer's positional embedding where both learned and fixed embeddings work well but the latter is preferred in practice \cite{vaswani2017attention}. For further experiments we use the [\textbf{ar + lin}] embedding. \Cref{tab:u-embed} shows the detailed effect of different embedding strategies.

\subsection{Distillation Losses and Proposal Distributions}
\label{sec:losses}

\Cref{eq:c-ptp-cross-entropy,eq:o-ptp-cross-entropy,eq:c-ptp-divergence} leave freedom as to how to choose the source of training sequences. In this section, we discuss and empirically compare several options.

If our goal is to deploy our parallelized student as a drop-in replacement of our teacher model, the lowest-variance option is to sample training sequences from the teacher.
Another possibility is to directly sample training sequences from a dataset, such as the one that was used to train the teacher model in the first place.
This has the advantage that we can compute the teacher predictions $Q_\varphi(t_k \mid t_{<k})$ in parallel over a full sequence instead of iteratively having to generate it.
Finally, we can sample sequences directly from the student model by first sampling auxiliary variables $u_i, \dots, u_T \sim \mathcal{U}[0, 1]$ and then using our student model in its current state to sample training sequences in parallel. As the student's prediction gets closer to that of the teacher during training, this approaches the same training sequence distribution as if we had sampled the teacher directly.

Conceptually, the latter is similar to the techniques used in distilling the autoregressive WaveNet into a parallel model \cite{oord2018parallel}. In practice, following their approach, which equates to using a reverse KL loss as noted above, is efficient but proved to be too unstable. Parallel WaveNet stabilizes training with several auxiliary losses, like a perceptual loss, which we do not have in our setting, in general.

We now empirically compare the distillation losses (\cref{sec:distillation}), focusing on KL and cross-entropy losses in \cref{eq:o-ptp-cross-entropy,eq:c-ptp-divergence}. Specifically the KL loss (\textbf{kl}), reverse KL loss (\textbf{kl-rev}), binary cross-entropy loss (\textbf{bce}), and categorical cross-entropy loss (\textbf{ce}). During training we sample training sequences from a dataset and continuations $t_{\geq i}$ either from the teacher model $Q_\varphi$, the student model $P_\theta$, or directly from a dataset. \Cref{tab:losses} shows the results for different losses. Empirically we note, that O-PTPs are easier to train than C-PTPs and achieve a higher number of average correct tokens. This is most likely due to the fact that O-PTPs do not have to predict the full token distribution accurately, which includes tail behavior, as long as they learn which token is the most likely given the auxiliary variable. In the following, we choose to sample training sequences from the teacher model for best results.

\begin{table}[]
    \centering
    \begin{tabular}{c|cc|cc|c}
        \toprule
        Proposal Distribution $P(t)$ & \textbf{kl} & \textbf{kl-rev} & \textbf{bce} & \textbf{ce} & MTP\\
        \midrule
        Teacher & $40$ & $41$ & $\mathbf{45}$ & $44$ & $10.1$ \\
        Student & $44$ & $39$ & $\mathbf{45}$ & $\mathbf{45}$ & $10.1$ \\
        Dataset & $29$ & $36$ & $44$ & $43$ & $10.1$ \\
        \bottomrule
    \end{tabular}
    \caption{\textbf{Our framework is compatible with several losses.} Average number of correct tokens ($\uparrow$) on the taxi dataset, evaluated on 16000 samples. O-PTP are distilled with KL or reverse KL loss (\textbf{kl}, \textbf{kl-rev}), C-PTP with binary or categorical cross entropy loss (\textbf{bce}, \textbf{ce}). Independent prediction (MTP) \cite{gloeckle2024better} achieves $10.1$. Numbers rounded to reflect level of statistical certainty.}
    \label{tab:losses}
\end{table}

\begin{table}
    \centering
    \begin{tabular}{c|ccccc|c}
        \toprule
        Model & \textbf{fl} + \textbf{NN} & \textbf{th} + \textbf{lin} & \textbf{th} + \textbf{NN} & \textbf{ar} + \textbf{lin} & \textbf{ar} + \textbf{NN} & MTP \\
        \midrule
        O-PTP & $35.9$ & $40.9$ & $39.1$ & $45.4$ & $\mathbf{46.1}$ & $10.1$\\
        C-PTP & $28.8$ & $36.8$ & $36.6$ & $\mathbf{40.4}$ & $35.7$ & $10.1$ \\
        \bottomrule
    \end{tabular}
    \caption{\textbf{Structured embeddings of auxiliary variables $u_k$ are more stable than fully-learned embeddings.} Average number of correct tokens ($\uparrow$) on the taxi dataset, evaluated on 16000 samples. Trained using the KL loss (C-PTP) and binary cross-entropy loss (O-PTP), respectively. Independent prediction (MTP) \cite{gloeckle2024better} achieves $10.1$. Numbers rounded to reflect level of statistical certainty.}
    \label{tab:u-embed}
\end{table}

\subsection{Sampling of Auxiliary Variables}
\label{sec:beta}

Our framework conditions, for a prompt $t_{<i}$, not on token $t_{k}$ directly but on the auxiliary variable $u_{k} \in [F_{k, t_{k}-1}, F_{k, t_{k}})$ that contains the same information. During inference we sample $u_k \sim \mathcal{U}[0, 1]$ as to not bias our predictions. During training on the other hand, we have more flexibility and can sample the permissible interval using 
$u_{k} = F_{k, t_{k}-1} + \tilde{u}_{k} \left[F_{k, t_{k}} - F_{k, t_{k}-1}\right]$, where $\tilde{u}_{k} \sim \mathrm{Beta}(b, b)$. For $b=1$ this simplifies to a uniform distribution while $b\neq1$ puts more or less weight on predictions that land closer to the border of the permissible interval and thus are more difficult to predict. Training results for different values of $b$ can be found in \Cref{tab:beta}. Empirically, we find that while the choice of $b$ does not seem to affect the final average number of correct samples, a larger $b$ might speed up the earlier stages of training while a smaller $b$ might yield slightly better sample quality during inference, as measured by model perplexity.

\begin{table}[]
    \centering
    \begin{tabular}{c|ccc|c}
        \toprule
        $b$ & $2$ & $1$ & $0.5$ & MTP \\
        \midrule
        O-PTP & $12.9$ & $\mathbf{13.9}$ & $13.8$ & $8.4$ \\
        C-PTP & $13.6$ & $\mathbf{13.8}$ & $13.5$ & $8.4$ \\
        \bottomrule
    \end{tabular}
    \caption{\textbf{Different sampling strategies for $u_k$ are available.} Average number of correct tokens ($\uparrow$) for $\tilde{u}_k \sim \mathrm{Beta}(b, b)$ on the taxi dataset, evaluated on 16000 samples, with $N=16$. Trained using the KL loss (C-PTP) and binary cross-entropy loss (O-PTP), respectively. Independent prediction (MTP) \cite{gloeckle2024better} achieves $8.4$. Numbers rounded to reflect level of statistical certainty.}
    \label{tab:beta}
\end{table}

\subsection{Abundant Computational Resources}
\label{sec:more_gpu}

Another way to leverage additional compute (via additional hardware or longer predictions) in a pure inference setting is to improve the expected number of correct tokens directly. Specifically, for a fixed context we can compute $M$ independent predictions using independently drawn auxiliary variables $u_{i,m}, \dots, u_{N, m}$ for $m = 1, \dots, M$. By choosing the best prediction, i.e. the one that gives us the best chance of a higher number of correct tokens, we can improve latency further.

Crucially, we have to choose the best prediction in a way that doesn't bias the marginal distribution over future tokens. If we, for example, naively choose the sequence that is correct for the most amount of tokens, we will bias our prediction towards sequences that are easier to predict. One way to achieve bias-free improvements is to pick the set of auxiliary variables that lands, on average, closest to the center of a token's valid interval $I_k(t_k) = [F_{k, t_{k}}, F_{k, t_{k}-1})$ where $F_{k, t_{k}}$ is the cumulative probability under $Q_\varphi$ to choose $t_{k}$ when predicting that token. Specifically, choose
\begin{align}
    \label{eq:maxu}
    \mathrm{argmax}_m \sum_{k=i}^{N}\abs{\frac{u_{k,m} - F_{k, t_{k,m}}}{F_{k, t_{k,m}-1} - F_{k, t_{k,m}}} - \frac{1}{2}}.
\end{align}
This does not bias the marginal distribution but does bias the distribution of the selected $u_k$ to be closer to the center of its interval $I_k(t_k)$, making the prediction less prone to small differences in the teacher's and student's logits. In the limit $M \to \infty$ we always select the middle point of $I_k(t_k)$, yielding an upper bound to the possible improvement. \Cref{tab:more_gpus} shows the performance gains on the taxi dataset.

\begin{table}[h]
    \centering
    \begin{tabular}{c|cccccccc}
        \toprule
        M & $1$ & $4$ & $16$ & $64$ & $256$ & $1024$ & $10^6$ & $\infty$ \\
        \midrule
        Avg. correct tokens & $45.36$ & $49.67$ & $54.76$ & $55.74$ & $56.91$ & $59.79$ & $65.42$ & $90.17$ \\
        \bottomrule
    \end{tabular}
    \caption{\textbf{Additional compute increases correctness}. Average number of correct tokens ($\uparrow$) for $M$ predicted sequences, computed in parallel on the taxi dataset, with sequence length $N=100$.}
    \label{tab:more_gpus}
\end{table}

\subsection{Restricted Computational Resources}
\label{sec:smalln}
Limiting the number $N$ of tokens our PTP predicts at once to a smaller number will reduce the total number of floating point operations, increasing energy efficiency. This, of course, negatively affects the possible latency gains, especially since $N$ is an upper bound on the average number of correct tokens. \Cref{tab:less_gpus} shows the result for different values of $N$ on the taxi dataset.

\begin{table}[h]
    \centering
    \begin{tabular}{c|cccccccc}
        \toprule
        N & $1$ & $2$ & $4$ & $8$ & $16$ & $64$ & $100$ & $\infty$ \\
        \midrule
        Avg. correct tokens & $1.00$ & $1.99$ & $3.93$ & $7.59$ & $13.90$ & $36.60$ & $45.36$ & $48.76$ \\
        MTP & $1.00$ & $1.91$ & $3.53$ & $5.86$ & $8.40$ & $10.08$ & $10.07$ & $10.20$ \\
        \bottomrule
    \end{tabular}
    \caption{\textbf{Less compute decreases correctness}. Average number of correct tokens ($\uparrow$) for limited number of predicted tokens $N$ per O-PTP call on the taxi dataset, $M=1$.}
    \label{tab:less_gpus}
\end{table}

\section{Efficient Error Correction}
\label{app:error-correction}

\Cref{alg:error-correction} shows how to correct errors with sequential calls to a proposal model and a base model.

\begin{figure}
    \centering
    \includegraphics[width=\linewidth]{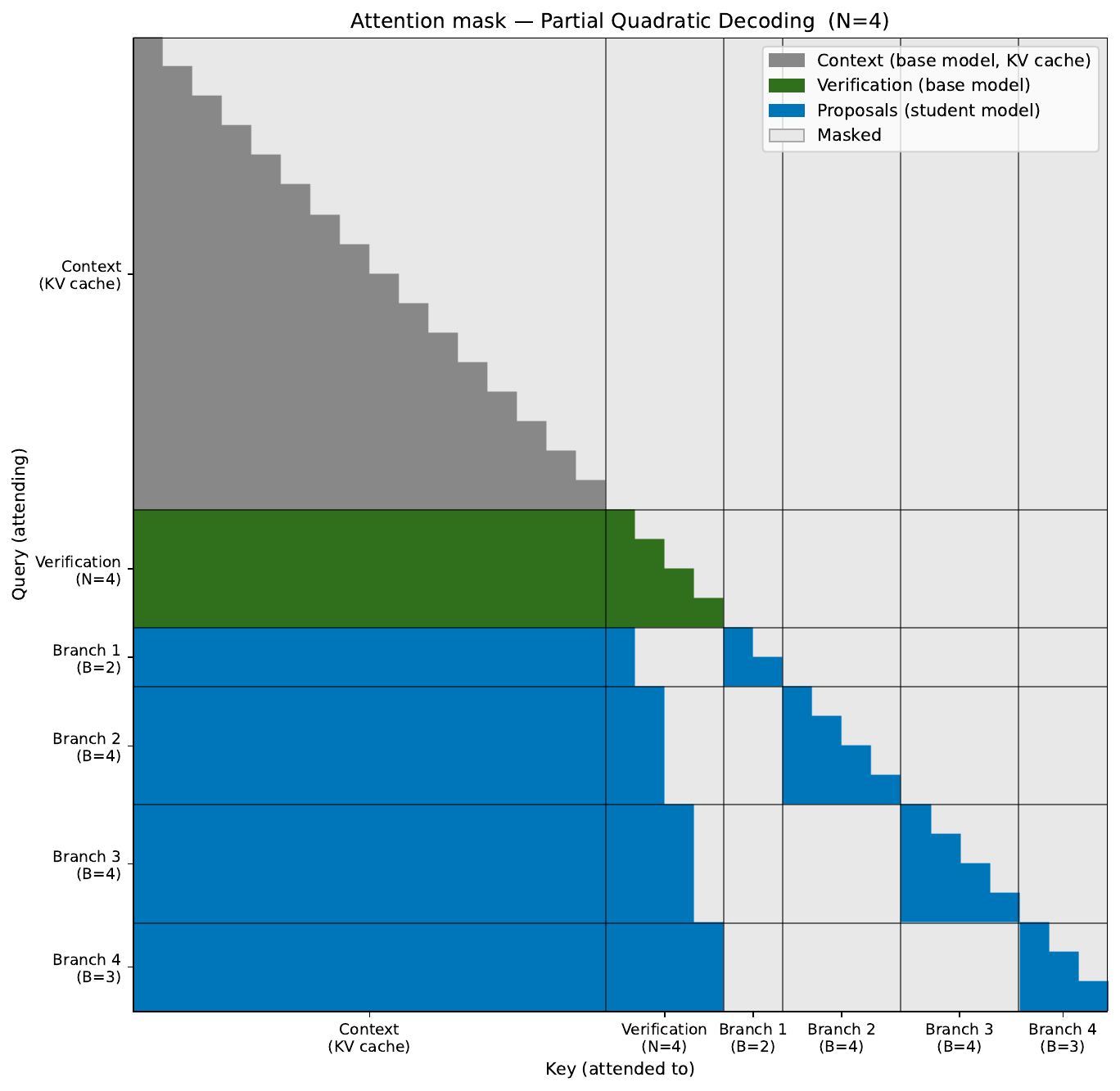}
    \caption{Attention mask used for efficient parallel decoding, when the PTP model is adapted from its base model via gated LoRA \cite{samragh2025your}. The attention mask corresponds to \cref{fig:error-correction}. The color in each row indicates what model weights are active: not active because contained in KV cache, base model for verification, and student model for proposals.}
    \label{fig:pqd-attention}
\end{figure}

\Cref{alg:error-correction-quadratic} shows how to correct errors by running a proposal model and a base model in parallel.
This can also be achieved in a single model call when leveraging Gated LoRA \cite{samragh2025your}: \Cref{alg:error-correction-quadratic} can be executed in a single forward pass by structuring the attention mask as shown in \cref{fig:pqd-attention}: the verification rows attend only to the prefix (base model weights active), while each proposal row attends to its hypothetical prefix plus all preceding proposal tokens (student weights active), enabling the base and student computations to run in parallel within one model call.

\begin{algorithm}[h]
\caption{Sampling with sequential error correction}
\label{alg:error-correction}
\begin{algorithmic}
\Require Prompt $t_{<i}$, one-hot or categorical PTP $P_\theta$, verification model $Q_\varphi$, target length $T$.
\State Sample $u_k \sim \Uu[0, 1]$ for all $k \geq i$.
\While{$i < T$}
\If{$P_\theta$ is one-hot PTP}
    \State $P_k \gets P_\theta(t_k \mid t_{<i}, u_i, \dots, u_{k})$, jointly for all $k \geq i$ \Comment{Student one-hot distributions}
    \State $t_k \gets \mathrm{argmax}_l P_{kl}$ for all $k \geq i$ \Comment{\cref{eq:o-ptp-sampling}}
\Else
    \State $P_k \gets P_\theta(t_k \mid t_{<i}, u_i, \dots, u_{k-1})$, jointly for all $k \geq i$ \Comment{Student categorical distributions}
    \State $t_k = \Pick(u_k, P_{k})$ for all $k \geq i$ \Comment{\cref{eq:c-ptp-sampling}}
\EndIf
\State $Q_k \gets Q_\varphi(t_k \mid t_{<k})$, jointly for all $k \geq i$ \Comment{Teacher categorical distributions}
\State $\tilde t_k \gets \Pick(u_k, Q_{k})$ for all $k \geq i$ \Comment{Teacher tokens}
\State $i \gets \min_{k \geq i} \{ k : t_k \neq \tilde t_k\}$ \Comment{cf.~\cref{eq:correct-count}}
\State $t_{i} \gets \tilde t_{i}$
\State $i \gets i + 1$ \Comment{$\accepted = \correct + 1$}
\EndWhile
\end{algorithmic}
\end{algorithm}

\begin{algorithm}[h]
\caption{Sampling with Partial Quadratic Decoding}
\label{alg:error-correction-quadratic}
\begin{algorithmic}
\Require Prompt $t_{<i}$, one-hot PTP $P_\theta$, verification model $Q_\varphi$, token-budget $C$, utility function $H$, target length $T$.
\State Sample $u_k \sim \Uu[0, 1]$ for all $k \geq i$.
\State $t_k \gets t_\mathrm{pad}$ for all $k \geq i$
\State $c_k \gets 0$ for all $k \geq i$ \Comment{Only allocate to $m=0$ in first iteration}
\While{$i < T$}
\State Compute optimal allocation $B$ of $C$ tokens from $c_k$ under $H$. \Comment{\cref{eq:opt-fixed}}
\State $P_{mk} \gets P_\theta(t_{k+m} \mid t_{<i+m}, u_{i+m}, \dots, u_{k+m})$, jointly for all $i + B_m > k \geq i, m \geq 0$
\State $t_{mk} \gets \mathrm{argmax}_l P_{mkl}$ \Comment{\cref{eq:o-ptp-sampling}}
\State $c_{mk} \gets \mathrm{max}_l P_{mkl}$ \Comment{Proposal Confidence}
\State $Q_k \gets Q_\varphi(t_k \mid t_{<k})$, jointly for all $k \geq i$ \Comment{$Q_\varphi$ and $P_\theta$ can be run in parallel}
\State $\tilde t_k \gets \Pick(u_k, Q_{k})$ for all $k \geq i$
\State $m \gets \min_{k \geq i} \{ k : t_k \neq \tilde t_k\} - i$ \Comment{cf.~\cref{eq:correct-count}}
\State $t_{k+m} \gets t_{mk}$ for $i + B_m > k \geq i$
\State $c_{k+m} \gets c_{mk}$ for $i + B_m > k \geq i$
\State $i \gets i + m$
\State $t_i \gets \tilde t_i$ 
\State $i \gets i + 1$ \Comment{$\accepted = \correct$}
\EndWhile
\end{algorithmic}
\end{algorithm}

\begin{figure}[h]
    \centering
    \includegraphics[width=\linewidth]{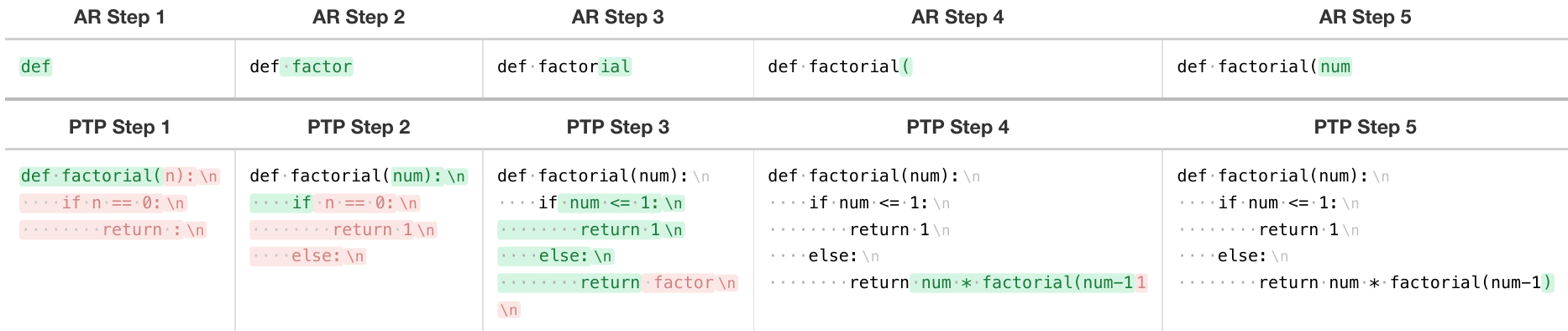}
    \caption{Detailed generation trajectory for \cref{fig:code-sampling}. Green is accepted, red is rejected by the base model call in the next step.}
    \label{fig:error-correction-example}
\end{figure}

\begin{figure}
    \centering
    \includegraphics[width=0.5\linewidth]{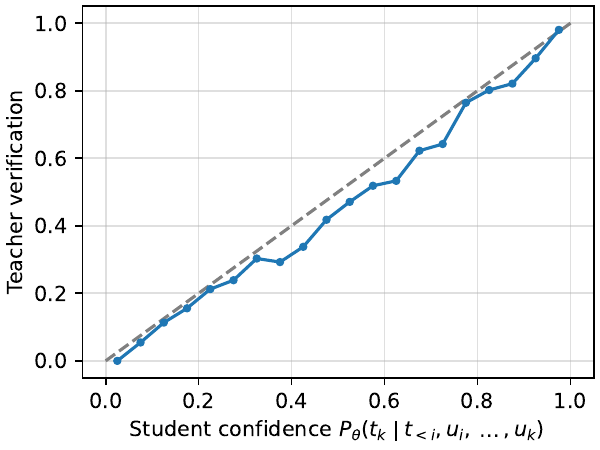}
    \caption{The learned probabilities $P_\theta(t_k \mid t_{<i}, u_i, \dots, u_k)$ of One-Hot Parallel Token Prediction (O-PTP) are reliable estimators as to whether a token will be accepted upon teacher verification.}
    \label{fig:o-ptp-calibration}
\end{figure}

\section{Experimental Details}
\label{sec:experimental-details}

\subsection{Training Algorithms}
\label{sec:algorithms}

\Cref{alg:ptp-distillation} shows how to distill a PTP from a teacher, \cref{alg:ptp-inverse-autoregressive} shows how to train directly from data.

\begin{algorithm}[h]
\caption{Training PTP (distillation)}
\label{alg:ptp-distillation}
\begin{algorithmic}
\Require Sequence proposal distribution $P(t)$ (teacher, student, dataset, or combination), cutoff distribution $P(i \mid t)$, teacher model $Q_\varphi(t)$, one-hot or categorical PTP $P_\theta$.
\While{not converged}
\State Sample $t \sim P(t)$
\State Sample $i \sim P(i \mid t)$.
\State Choose $N$ such that $i < N \leq T$.
\State $P_k = Q_\varphi(t_k \mid t_{<k})$ in single model call for all $k=i, \dots, N$.
\State Sample $u_k \in [F_{k,t_k-1}, F_{k,t_k})$ for all $k=i, \dots, N$.
\State Compute $\nabla_\theta \mathcal{L}(\theta, t)$ using one of \cref{eq:o-ptp-cross-entropy,eq:c-ptp-cross-entropy,eq:c-ptp-divergence}.
\State Gradient step.
\EndWhile
\end{algorithmic}
\end{algorithm}

\begin{algorithm}[h]
\caption{Training PTP (inverse autoregressive)}
\label{alg:ptp-inverse-autoregressive}
\begin{algorithmic}
\Require Dataset $P(t)$, cutoff distribution $P(i \mid t)$, categorical PTP $P_\theta$.
\While{not converged}
\State Sample training sequence $t_1, \dots, t_T$.
\State Sample split position $i \in \{1, \dots, T\}$.
\State Choose $N$ such that $i < N \leq T$.
\For{$k = i, \dots, N$}
\State $P_k = P_\theta(t_k \mid t_{<i}, u_{i}, \dots, u_{k-1})$, with auxiliary available from previous iterations.
\State Sample $u_k \in [F_{k,t_k-1}, F_{k,t_k})$.
\EndFor
\State Compute $\nabla_\theta \mathcal{L}(\theta, t)$ using \cref{eq:ce-iat}.
\State Gradient step.
\EndWhile
\end{algorithmic}
\end{algorithm}

\subsection{Training Details}
\label{sec:training-details}

The model used in \cref{sec:inverse-autoregressive-experiment,app:ablation} is a GPT-2–style transformer language model with 4 transformer layers, a hidden size of 1536, and approximately 29 million trainable parameters. Each layer follows the standard GPT-2 architecture, consisting of multi-head self-attention and position-wise feedforward sublayers, combined with residual connections and layer normalization. The vocabulary size is set to $25$. Unless otherwise noted, all other hyperparameters and initialization schemes follow the original GPT-2 specification \cite{radford2019language}. During training and inference of our student model we don't provide any context and evaluate the correctness of the next $N=100$ tokens, by comparing $Q_\varphi(t_k \mid t_{<k})$ and $P_\theta(t_k)$. For results on a smaller $N=16$, see \cref{sec:smalln}. We train every model for $150$k steps with a batch size of $32$ with the Adam optimizer \cite{kingma2015adam} and learning rate $0.0001$.

The teacher model used in \cref{sec:competition-limitations} is a dialogue-tuned variant of the TinyLlama \cite{zhang2024tinyllama} 1.1 billion parameter model, adopting the same architecture and tokenizer as LLaMA 2 \cite{touvron2023llama2}: \texttt{TinyLlama-1.1B-Chat-v1.0}. The model uses a transformer architecture comprising $22$ transformer layers, each with standard multi-head self-attention, SwiGLU feedforward blocks, residual connections, and layer normalization. The embedding and hidden dimension is $2048$, and the intermediate (feedforward) dimension is $5632$, consistent with a LLaMA-style scaling. The vocabulary size is $32,000$. The parameters are available via \url{https://huggingface.co/TinyLlama/TinyLlama-1.1B-Chat-v1.0}.

We train an O-PTP on predicting 16 additional tokens. We adopt the strategy of finetuning using gated low-rank adaptation, where changed parameters are only applied for completion tokens~\mbox{\cite{samragh2025your}}. We choose a LoRA \cite{hu2022lora} rank of $r=256$, although comparable speedups can be achieved already with $r=64$. We train every model for 1.3M steps with a batch size of $56$ with the AdamW optimizer \cite{loshchilov2019decoupled} on \cref{eq:o-ptp-cross-entropy} and learning rate $0.0001$. We generate training and validation data by generating Python code completions on CodeContests \cite{li2022competitionlevel} from the teacher, splitting generated sequences randomly into input and completion. We use a teacher sampling temperature of 0.7, top-$k=50$ and top-$p=0.9$, as is recommended for this model.

For the MTP baseline, we use \cref{eq:c-ptp-cross-entropy} with uninformative $u$s in otherwise identical code for a fair comparison.

For the scaling experiments in \cref{sec:speculative-decoding}, we parameterize models of different scales as in \cref{tab:scaling-model-configs}. We adopt the same architecture choices as TinyLlama.

For the natural language tasks on SpecBench \cite{xia2024unlocking}, we finetune Vicuna-7B-v1.5~\cite{vicuna2023} on ShareGPT \cite{chen2024sharegpt4v}. We train it with the same gated $r=128$ LoRA setup as for TinyLlama, but with batch size $64$ and for 1M steps.

\begin{table}[h]
\centering
\begin{tabular}{lccc}
\toprule
Model Size & Hidden Size & Hidden Layers & Attention Heads \\
\midrule
4k    & 16   & 1  & 1  \\
66k   & 64   & 1  & 1  \\
525k  & 128  & 2  & 2  \\
4.2M  & 256  & 4  & 4  \\
34M   & 512  & 8  & 8  \\
268M  & 1024 & 16 & 16 \\
\midrule
1.1B (teacher)  & 2048 & 22 & 32 \\
\bottomrule
\end{tabular}
\caption{Model configuration per parameter count for scaling experiment}
\label{tab:scaling-model-configs}
\end{table}

We base our code on \texttt{transformers} \cite{wolf2019huggingface}, \texttt{PyTorch} \cite{paszke2019pytorch}, PyTorch Lightning \citep{falcon2019pytorch}, Numpy \citep{harris2020array}, Matplotlib \citep{hunter2007matplotlib} for plotting and Pandas \citep{mckinney2010data} for data evaluation.

\section{Self-Verification}
\label{app:self-verification}

The experiments in \cref{sec:vicuna} use a separate teacher model for verification, guaranteeing output identical in distribution to the teacher.
Here we explore a faster alternative that requires no teacher at all: the PTP student model verifies its own proposals.

After drafting tokens with auxiliary variables $u_k$, the student evaluates its marginal distribution $P_\theta(\cdot \mid t_{<k})$ with the gated adapter weights active but \emph{without} any auxiliary variable input.
Per \cref{thm:C-PTP}, this recovers the standard autoregressive distribution of the student.
A proposed token $t_k$ is accepted if it matches $\Pick(u_k,\, P_\theta(t_k \mid t_{<k}))$.
Since both proposal and verification use the same model in a single forward pass, no separate teacher call is required.

Under self-verification on SpecBench, the average number of accepted tokens is $\accepted{} = 6.1$, compared to $\accepted{} = 4.2$ under teacher verification (\cref{tab:spec-bench}).
The higher acceptance rate arises because the student's drafts are already well-aligned with its own marginal distribution.
This translates to a significantly higher potential speedup than what is reported in the main text, at the cost of no longer guaranteeing identity with the teacher's output distribution.

Below, we show uncurated outputs from SpecBench prompts \cite{xia2024unlocking}.
Despite the absence of teacher verification, the student produces coherent, instruction-following responses.

\subsection{Coding}
\textbf{Prompt:} Write a C++ program to find the nth Fibonacci number using recursion.

\noindent\textbf{Response:}
\begin{lstlisting}
Here is a simple program that finds the nth Fibonacci number using recursion in C++:
```
#include <iostream>

using namespace std;

int fibonacci(int n) {
    if (n <= 0) {
        return 0;
    } else if (n == 1) {
        return 1;
    } else if (n == 2) {
        return 2;
    } else {
        return fibonacci(n-1) + fibonacci(n-2);
    }
}

int main() {
    int n;
    cout << "Enter the number of Fibonacci numbers to find: ";
    cin >> n;
    cout << "The first " << n << " Fibonacci numbers are:\n";
    for (int i = 0; i < n; i++) {
        cout << "Fibonacci(" << i << ") = " << fibonacci(i) << endl;
    }
    return 0;
}
```
This program asks the user for the number of Fibonacci numbers they want to find, and then prints out the first n Fibonacci numbers. The function `fibonacci` is called recursively to compute the nth Fibonacci number, with the base cases being 0, 1, and 2. The `main` function is used to handle the user input and print out the results.

Note: This is a very simple implementation of the Fibonacci sequence, and it may not be suitable for larger values of n. For larger values of n, you may need to use a different algorithm, such as the Golden Ratio algorithm or the Matrix Exponential algorithm. Additionally, this implementation assumes that the input value is positive, so you may need to handle negative input values appropriately.
\end{lstlisting}

\subsection{Extraction}
\textbf{Prompt:} Evaluate the following movie reviews on a scale of 1 to 5, with 1 being very negative, 3 being neutral, and 5 being very positive:
1. This movie released on Nov. 18, 2019, was phenomenal. The cinematography, the acting, the plot - everything was top-notch.
2. Never before have I been so disappointed with a movie. The plot was predictable and the characters were one-dimensional. In my opinion, this movie is the worst one to have been released in 2022.
3. The movie was okay. There were some parts I  enjoyed, but there were also parts that felt lackluster. This is a movie that was released in Feb 2018 and seems to be quite ordinary.
Return the answer as a JSON array of integers.

\noindent\textbf{Response:}
\begin{lstlisting}
[1, 4, 3]
\end{lstlisting}

\subsection{Math}
\textbf{Prompt:} When rolling two dice, what is the probability that you roll a total number that is at least 3?

\noindent\textbf{Response:}
\begin{lstlisting}
There are two ways to roll a total number that is at least 3, 3 and 3, or 4 and 2. So the probability is 1/36.

To calculate this, you can use the formula for combinations, which is:

n(n+1)/2 = P(A)

where n is the number of choices, and P(A) is the probability of A occurring. In this case, n = 2, so the probability of rolling a total number that is at least 3 is 1/2\*1/2 = 1/36.

Alternatively, you can use the formula for the binomial distribution, which is:

P(A) = (n! / (n-k)!) \* (1-p)^k \* p^n

where n is the number of trials, k is the number of successes, and p is the probability of success. In this case, n = 2, k = 1, and p = 1/36. Plugging these values into the formula, we get P(A) = (2! / (2-1)!) \* (1-1/36)^1 \* 36/36 = 1/36.

I hope this helps! Let me know if you have any more questions.
\end{lstlisting}

\subsection{Math Reasoning}
\textbf{Prompt:} A company wanted to buy 500 computers and had \$700 for each computer. However, when they went to order, they realized the price of each computer was 10\% higher than they had planned for. Calculate the total amount of money they paid to buy the computers at the new prices.

\noindent\textbf{Response:}
\begin{lstlisting}
The price of each computer was 10%
The company wanted to buy 500 computers, so they paid $770 \* 500 = $38500 for the computers.
Therefore, the company paid $38,500 to buy the computers at the new prices.
It's important to note that this is just an example, and the actual amount of money paid by the company would depend on the specifics of the purchase, such as any discounts or taxes that may have been applied. The calculation provided here is just a general example of how to calculate the total amount of money paid for the computers at the new prices.
\end{lstlisting}

\subsection{QA}
\textbf{Prompt:} Name of the last part of harry potter?

\noindent\textbf{Response:}
\begin{lstlisting}
The last part of Harry Potter is "Harry Potter and the Deathly Hallows, Part 2." It was released in 2007 and is the final installment in the Harry Potter series by J.K. Rowling. The film adaptation of the book was directed by David Yates and starred Daniel Radcliffe, Emma Watson, and Rupert Grint. The Deathly Hallows Part 2 follows the story of Harry Potter and his friends as they confront the dark lord Voldemort and the Deathly Hallows, a powerful and dangerous object that grants its possessor immense power. The film was a critical and commercial success, and it brought an end to the Harry Potter series, which had captivated audiences worldwide for over a decade. The series as a whole is considered one of the most successful and influential book and film series in recent history, and its legacy continues to be felt in popular culture.
\end{lstlisting}

\subsection{Reasoning}
\textbf{Prompt:} One morning after sunrise, Suresh was standing facing a pole. The shadow of the pole fell exactly to his right. Can you tell me the direction towards which the shadow was pointing - east, south, west, or north? Explain your reasoning steps.

\noindent\textbf{Response:}
\begin{lstlisting}
To determine the direction towards which the shadow was pointing, we need to use the relationship between the angle of the shadow and the angle of the sunrise.

Firstly, we know that the sun rises in the east and sets in the west. So, if the shadow is pointing towards the east, then the pole is pointing towards the west.

Now, we need to determine the angle between the shadow and the pole. Since the shadow is pointing towards the east, we can draw an arrow from the shadow to the east and another arrow from the pole to the west.

The angle between these two arrows will give us the angle of the shadow. Since the shadow is pointing towards the east, the angle between the arrow and the pole will be a right angle, or 90 degrees.

Therefore, the direction towards which the shadow was pointing was east.

In summary, the pole was pointing towards the east, and the shadow was pointing towards the east as well. The angle of the shadow was 90 degrees, which means the pole and the shadow were in line with each other. So, the direction towards which the shadow was pointing was east.
\end{lstlisting}

\subsection{Roleplay}
\textbf{Prompt:} Embrace the role of Sheldon from "The Big Bang Theory" as we delve into our conversation. Don't start with phrases like "As Sheldon". Let's kick things off with the following question: "What is your opinion on hand dryers?"

\noindent\textbf{Response:}
\begin{lstlisting}
As Sheldon, I have a bit of a love-hate relationship with hand dryers. On one hand, I find them to be quite efficient and convenient, especially when compared to paper towels or cloth towels. However, on the other hand, I feel that they can be quite loud and disruptive, especially in public restrooms. Overall, I think they are a decent choice for drying your hands, but it's important to be mindful of others around you and try not to use them in situations where they may be a nuisance.
\end{lstlisting}

\subsection{Stem}
\textbf{Prompt:} How have the Alps and Rhine River influenced settlement and agriculture in Western Europe? List three impacts.

\noindent\textbf{Response:}
\begin{lstlisting}
The Alps and Rhine River have had a significant impact on settlement and agriculture in Western Europe. Here are three ways in which this has occurred:

1. Climate: The Alps act as a natural barrier, blocking the flow of warm air and creating a cooler climate in the region. This has made it difficult to grow certain crops, such as wheat, which requires a warmer climate. However, the cooler climate has also made it easier to grow other crops, such as fruit and vegetables, which thrive in cooler temperatures.
2. Soil: The Rhine River has created a fertile soil deposit along its banks, making it an ideal location for agriculture. The river also provides a source of water for irrigation, which helps to support the growth of crops.
3. Trade: The Rhine River has been an important trade route for centuries, connecting Western Europe with other parts of the continent. This has made it easier for farmers in the region to sell their crops and has helped to support the growth of settlements along the river. The Alps, on the other hand, have created natural barriers that have made it more difficult to travel and trade goods. However, the presence of the mountains has also made it easier for farmers to protect their crops from invasions and other threats.

Overall, the Alps and Rhine River have had a significant impact on settlement and agriculture in Western Europe, shaping the region's climate, soil, and trade routes, and influencing the types of crops that can be grown and the settlements that can be built. These factors have contributed to the development of distinctive agricultural and settlement patterns in the region.
\end{lstlisting}

\subsection{Translation}
\textbf{Prompt:} Translate German to English: Nicht zu vergessen die richtige Kosmetik und Nagelpflege .

\noindent\textbf{Response:}
\begin{lstlisting}
Don't forget about the right cosmetics and nail care.
\end{lstlisting}

\subsection{Writing}
\textbf{Prompt:} Compose an engaging travel blog post about a recent trip to Hawaii, highlighting cultural experiences and must-see attractions.

\noindent\textbf{Response:}
\begin{lstlisting}
Title: A Hawaii Adventure: Exploring the Islands' Rich Culture and Natural Beauty

Introduction:
Hawaii is a tropical paradise that has captivated the hearts of many travelers with its stunning natural wonders, crystal-clear waters, and lush landscapes. But what sets Hawaii apart from other vacation spots is its rich culture and history. In this blog post, we'll take a journey through the islands' cultural experiences and must-see attractions, leaving you with a deeper appreciation for the land of the setting sun.

Day 1: Oahu

* Visited the USS Arizona Memorial at Pearl Harbor, a solemn reminder of the tragic event that led to World War II.
* Explored the Nuuanu Pali Trail, offering breathtaking views of the island and the surrounding ocean.
* Enjoyed a traditional Hawaiian luau dinner, where we learned about Hawaiian culture, history, and traditions.
Day 2: Maui

* Visited the Hanauma Bay Nature Preserve, a stunning marine sanctuary where we could swim and snorkel among colorful fish and vibrant coral reefs.
* Explored the iconic Road to Hana, stopping at picturesque viewpoints and waterfalls along the way.
* Went whale watching and witnessed the majestic creatures in their natural habitat.

Day 3: Kauai

* Explored the Napali Coastline, a stunning series of cliffs, waterfalls, and beaches that offer breathtaking views of the ocean.
* Visited the Hanalei Bay, where we could swim, sunbathe on the beach, and watch the sunset.
* Enjoyed a sunset sail, where we could watch the sunset while enjoying the cool ocean breeze.
Day 4: Big Island

* Visited the Hawaii Volcanoes National Park, where we could see the Kilauea Volcano and explore the park's lush rainforests and waterfalls.
* Visited the Akaka Falls State Park, where we could see the stunning waterfall and its surrounding natural beauty.
* Enjoyed a farm tour and tasting at a local organic farm, where we could learn about sustainable farming practices and enjoy delicious fresh produce.

Conclusion:
Hawaii is more than just a vacation spot, it's a cultural and historical adventure. From the USS Arizona Memorial to the Akaka Falls State Park, there is something for everyone in Hawaii. Whether you're a nature lover or a cultural enthusiast, Hawaii has something special to offer. So pack your bags, grab your camera, and get ready for an adventure that you'll never forget.
\end{lstlisting}

\end{document}

%% file: math_commands.tex
\usepackage{amsmath,amsfonts,bm}

\def\eqref#1{equation~\ref{#1}}

\def\1{\bm{1}}

\DeclareMathAlphabet{\mathsfit}{\encodingdefault}{\sfdefault}{m}{sl}
\SetMathAlphabet{\mathsfit}{bold}{\encodingdefault}{\sfdefault}{bx}{n}

%% file: iclr2026_conference.bib
@inproceedings{gu2018nonautoregressive,
    title = {Non-autoregressive neural machine translation},
    url = {https://openreview.net/forum?id=B1l8BtlCb},
    booktitle = {International conference on learning representations},
    author = {Gu, Jiatao and Bradbury, James and Xiong, Caiming and Li, Victor O.K. and Socher, Richard},
    year = {2018},
}

@inproceedings{leviathan2023fast,
    series = {Proceedings of machine learning research},
    title = {Fast inference from transformers via speculative decoding},
    volume = {202},
    url = {https://proceedings.mlr.press/v202/leviathan23a.html},
    booktitle = {Proceedings of the 40th international conference on machine learning},
    publisher = {PMLR},
    author = {Leviathan, Yaniv and Kalman, Matan and Matias, Yossi},
    month = jul,
    year = {2023},
    pages = {19274--19286},
}

@inproceedings{sun2023spectr,
    title = {{SpecTr}: {Fast} speculative decoding via optimal transport},
    volume = {36},
    url = {https://proceedings.neurips.cc/paper_files/paper/2023/file/6034a661584af6c28fd97a6f23e56c0a-Paper-Conference.pdf},
    booktitle = {Advances in neural information processing systems},
    publisher = {Curran Associates, Inc.},
    author = {Sun, Ziteng and Suresh, Ananda Theertha and Ro, Jae Hun and Beirami, Ahmad and Jain, Himanshu and Yu, Felix},
    editor = {Oh, A. and Naumann, T. and Globerson, A. and Saenko, K. and Hardt, M. and Levine, S.},
    year = {2023},
    pages = {30222--30242},
}

@inproceedings{zhong2025speeding,
    title = {Speeding up speculative decoding via sequential approximate verification},
    url = {https://openreview.net/forum?id=Y4KcfotBkf},
    booktitle = {{ES}-{FoMo} {III}: 3rd workshop on efficient systems for foundation models},
    author = {Zhong, Meiyu and Teku, Noel and Tandon, Ravi},
    year = {2025},
}

@article{chen2023accelerating,
    title = {Accelerating large language model decoding with speculative sampling},
    journal = {arXiv preprint arXiv:2302.01318},
    author = {Chen, Charlie and Borgeaud, Sebastian and Irving, Geoffrey and Lespiau, Jean-Baptiste and Sifre, Laurent and Jumper, John},
    year = {2023},
}

@inproceedings{ma2019flowseq,
    address = {Hong Kong},
    title = {{FlowSeq}: {Non}-autoregressive conditional sequence generation with generative flow},
    booktitle = {Proceedings of the 2019 conference on empirical methods in natural language processing},
    author = {Ma, Xuezhe and Zhou, Chunting and Li, Xian and Neubig, Graham and Hovy, Eduard},
    month = nov,
    year = {2019},
}

@inproceedings{gloeckle2024better,
    title = {Better \& faster large language models via multi-token prediction},
    booktitle = {Proceedings of the 41st international conference on machine learning},
    author = {Gloeckle, Fabian and Idrissi, Badr Youbi and Rozière, Baptiste and Lopez-Paz, David and Synnaeve, Gabriel},
    year = {2024},
    pages = {15706--15734},
}

@inproceedings{qi2020prophetnet,
    address = {Online},
    title = {{ProphetNet}: {Predicting} future n-gram for sequence-to-{SequencePre}-training},
    url = {https://aclanthology.org/2020.findings-emnlp.217/},
    doi = {10.18653/v1/2020.findings-emnlp.217},
    booktitle = {Findings of the association for computational linguistics: {EMNLP} 2020},
    publisher = {Association for Computational Linguistics},
    author = {Qi, Weizhen and Yan, Yu and Gong, Yeyun and Liu, Dayiheng and Duan, Nan and Chen, Jiusheng and Zhang, Ruofei and Zhou, Ming},
    editor = {Cohn, Trevor and He, Yulan and Liu, Yang},
    month = nov,
    year = {2020},
    pages = {2401--2410},
}

@misc{deepseek-ai2025deepseekv3,
    title = {{DeepSeek}-{V3} {Technical} {Report}},
    url = {http://arxiv.org/abs/2412.19437},
    doi = {10.48550/arXiv.2412.19437},
    urldate = {2025-08-16},
    publisher = {arXiv},
    author = {DeepSeek-AI and Liu, Aixin and Feng, Bei and Xue, Bing and Wang, Bingxuan and Wu, Bochao and Lu, Chengda and Zhao, Chenggang and Deng, Chengqi and Zhang, Chenyu and Ruan, Chong and Dai, Damai and Guo, Daya and Yang, Dejian and Chen, Deli and Ji, Dongjie and Li, Erhang and Lin, Fangyun and Dai, Fucong and Luo, Fuli and Hao, Guangbo and Chen, Guanting and Li, Guowei and Zhang, H. and Bao, Han and Xu, Hanwei and Wang, Haocheng and Zhang, Haowei and Ding, Honghui and Xin, Huajian and Gao, Huazuo and Li, Hui and Qu, Hui and Cai, J. L. and Liang, Jian and Guo, Jianzhong and Ni, Jiaqi and Li, Jiashi and Wang, Jiawei and Chen, Jin and Chen, Jingchang and Yuan, Jingyang and Qiu, Junjie and Li, Junlong and Song, Junxiao and Dong, Kai and Hu, Kai and Gao, Kaige and Guan, Kang and Huang, Kexin and Yu, Kuai and Wang, Lean and Zhang, Lecong and Xu, Lei and Xia, Leyi and Zhao, Liang and Wang, Litong and Zhang, Liyue and Li, Meng and Wang, Miaojun and Zhang, Mingchuan and Zhang, Minghua and Tang, Minghui and Li, Mingming and Tian, Ning and Huang, Panpan and Wang, Peiyi and Zhang, Peng and Wang, Qiancheng and Zhu, Qihao and Chen, Qinyu and Du, Qiushi and Chen, R. J. and Jin, R. L. and Ge, Ruiqi and Zhang, Ruisong and Pan, Ruizhe and Wang, Runji and Xu, Runxin and Zhang, Ruoyu and Chen, Ruyi and Li, S. S. and Lu, Shanghao and Zhou, Shangyan and Chen, Shanhuang and Wu, Shaoqing and Ye, Shengfeng and Ye, Shengfeng and Ma, Shirong and Wang, Shiyu and Zhou, Shuang and Yu, Shuiping and Zhou, Shunfeng and Pan, Shuting and Wang, T. and Yun, Tao and Pei, Tian and Sun, Tianyu and Xiao, W. L. and Zeng, Wangding and Zhao, Wanjia and An, Wei and Liu, Wen and Liang, Wenfeng and Gao, Wenjun and Yu, Wenqin and Zhang, Wentao and Li, X. Q. and Jin, Xiangyue and Wang, Xianzu and Bi, Xiao and Liu, Xiaodong and Wang, Xiaohan and Shen, Xiaojin and Chen, Xiaokang and Zhang, Xiaokang and Chen, Xiaosha and Nie, Xiaotao and Sun, Xiaowen and Wang, Xiaoxiang and Cheng, Xin and Liu, Xin and Xie, Xin and Liu, Xingchao and Yu, Xingkai and Song, Xinnan and Shan, Xinxia and Zhou, Xinyi and Yang, Xinyu and Li, Xinyuan and Su, Xuecheng and Lin, Xuheng and Li, Y. K. and Wang, Y. Q. and Wei, Y. X. and Zhu, Y. X. and Zhang, Yang and Xu, Yanhong and Xu, Yanhong and Huang, Yanping and Li, Yao and Zhao, Yao and Sun, Yaofeng and Li, Yaohui and Wang, Yaohui and Yu, Yi and Zheng, Yi and Zhang, Yichao and Shi, Yifan and Xiong, Yiliang and He, Ying and Tang, Ying and Piao, Yishi and Wang, Yisong and Tan, Yixuan and Ma, Yiyang and Liu, Yiyuan and Guo, Yongqiang and Wu, Yu and Ou, Yuan and Zhu, Yuchen and Wang, Yuduan and Gong, Yue and Zou, Yuheng and He, Yujia and Zha, Yukun and Xiong, Yunfan and Ma, Yunxian and Yan, Yuting and Luo, Yuxiang and You, Yuxiang and Liu, Yuxuan and Zhou, Yuyang and Wu, Z. F. and Ren, Z. Z. and Ren, Zehui and Sha, Zhangli and Fu, Zhe and Xu, Zhean and Huang, Zhen and Zhang, Zhen and Xie, Zhenda and Zhang, Zhengyan and Hao, Zhewen and Gou, Zhibin and Ma, Zhicheng and Yan, Zhigang and Shao, Zhihong and Xu, Zhipeng and Wu, Zhiyu and Zhang, Zhongyu and Li, Zhuoshu and Gu, Zihui and Zhu, Zijia and Liu, Zijun and Li, Zilin and Xie, Ziwei and Song, Ziyang and Gao, Ziyi and Pan, Zizheng},
    month = feb,
    year = {2025},
}

@article{radford2018improving,
  title={Improving language understanding by generative pre-training},
  author={Radford, Alec and Narasimhan, Karthik and Salimans, Tim and Sutskever, Ilya and others},
  journal={{OpenAI}},
  year={2018},
  publisher={San Francisco, CA, USA}
}

@article{vaswani2017attention,
  title={Attention is all you need},
  author={Vaswani, Ashish and Shazeer, Noam and Parmar, Niki and Uszkoreit, Jakob and Jones, Llion and Gomez, Aidan N and Kaiser, {\L}ukasz and Polosukhin, Illia},
  journal={Advances in neural information processing systems},
  volume={30},
  year={2017}
}

@inproceedings{brown2020language,
    title = {Language models are few-shot learners},
    volume = {33},
    url = {https://proceedings.neurips.cc/paper_files/paper/2020/file/1457c0d6bfcb4967418bfb8ac142f64a-Paper.pdf},
    booktitle = {Advances in neural information processing systems},
    publisher = {Curran Associates, Inc.},
    author = {Brown, Tom and Mann, Benjamin and Ryder, Nick and Subbiah, Melanie and Kaplan, Jared D and Dhariwal, Prafulla and Neelakantan, Arvind and Shyam, Pranav and Sastry, Girish and Askell, Amanda and Agarwal, Sandhini and Herbert-Voss, Ariel and Krueger, Gretchen and Henighan, Tom and Child, Rewon and Ramesh, Aditya and Ziegler, Daniel and Wu, Jeffrey and Winter, Clemens and Hesse, Chris and Chen, Mark and Sigler, Eric and Litwin, Mateusz and Gray, Scott and Chess, Benjamin and Clark, Jack and Berner, Christopher and McCandlish, Sam and Radford, Alec and Sutskever, Ilya and Amodei, Dario},
    editor = {Larochelle, H. and Ranzato, M. and Hadsell, R. and Balcan, M.F. and Lin, H.},
    year = {2020},
    pages = {1877--1901},
}

@misc{zhang2024tinyllama,
    title = {{TinyLlama}: {An} open-source small language model},
    author = {Zhang, Peiyuan and Zeng, Guangtao and Wang, Tianduo and Lu, Wei},
    year = {2024},
    note = {arXiv: 2401.02385 [cs.CL]},
}

@article{li2023starcoder,
    title = {{StarCoder}: {May} the source be with you!},
    journal = {Transactions on machine learning research},
    author = {Li, R and Allal, LB and Zi, Y and Muennighoff, N and Kocetkov, D and Mou, C and Marone, M and Akiki, C and Li, J and Chim, J and {others}},
    year = {2023},
    note = {Publisher: OpenReview},
}

@article{li2022competitionlevel,
    title = {Competition-level code generation with {AlphaCode}},
    volume = {378},
    url = {https://www.science.org/doi/abs/10.1126/science.abq1158},
    doi = {10.1126/science.abq1158},
    number = {6624},
    journal = {Science},
    author = {Li, Yujia and Choi, David and Chung, Junyoung and Kushman, Nate and Schrittwieser, Julian and Leblond, Rémi and Eccles, Tom and Keeling, James and Gimeno, Felix and Lago, Agustin Dal and Hubert, Thomas and Choy, Peter and de Masson d’Autume, Cyprien and Babuschkin, Igor and Chen, Xinyun and Huang, Po-Sen and Welbl, Johannes and Gowal, Sven and Cherepanov, Alexey and Molloy, James and Mankowitz, Daniel J. and Robson, Esme Sutherland and Kohli, Pushmeet and de Freitas, Nando and Kavukcuoglu, Koray and Vinyals, Oriol},
    year = {2022},
    pages = {1092--1097},
}

@inproceedings{guo2017calibration,
    title = {On calibration of modern neural networks},
    booktitle = {International conference on machine learning},
    publisher = {PMLR},
    author = {Guo, Chuan and Pleiss, Geoff and Sun, Yu and Weinberger, Kilian Q},
    year = {2017},
    pages = {1321--1330},
}

@inproceedings{holtzman2020curious,
    title = {The curious case of neural text degeneration},
    booktitle = {International conference on learning representations},
    author = {Holtzman, Ari and Buys, Jan and Du, Li and Forbes, Maxwell and Choi, Yejin},
    year = {2020},
}

@misc{nyc_tlc_2016,
    author       = {{New York City Taxi and Limousine Commission}},
    shortauthor   = {{NYC TLC}},
    title        = {2016 Yellow Taxi Trip Data},
    year         = {2017},
    note         = {City of New York, OpenData portal},
}

@article{radford2019language,
  title={Language Models are Unsupervised Multitask Learners},
  author={Radford, Alec and Wu, Jeffrey and Child, Rewon and Luan, David and Amodei, Dario and Sutskever, Ilya},
  journal={OpenAI Technical Report},
  year={2019}
}

@article{kingma2015adam,
  title={Adam: A Method for Stochastic Optimization},
  author={Kingma, Diederik P. and Ba, Jimmy},
  journal={International Conference on Learning Representations (ICLR)},
  year={2015},
}

@inproceedings{loshchilov2019decoupled,
  title={Decoupled Weight Decay Regularization},
  author={Loshchilov, Ilya and Hutter, Frank},
  booktitle={International Conference on Learning Representations (ICLR)},
  year={2019},
}

@book{witten1987arithmetic,
  title={Arithmetic Coding for Data Compression},
  author={Witten, Ian H. and Neal, Radford M. and Cleary, John G.},
  year={1987},
  publisher={Communications of the ACM},
  volume={30},
  pages={520--540},
}

@article{touvron2023llama2,
  title     = {LLaMA 2: Open Foundation and Fine-Tuned Chat Models},
  author    = {Touvron, Hugo and Martin, Louis and Stone, Kevin and Albert, Peter and Almahairi, Amjad and Babaei, Yasmine and Bashlykov, Nikolay and Batra, Soumya and Bhargava, Prajjwal and Bhosale, Shruti and Bikel, Daniel and Blecher, Lukas and Bogoychev, Nikolay and Brannon, William and Brohan, Anthony and Caballero, Humberto and Chadwick, Andy and Lee, Jenny and others},
  year      = {2023},
  journal   = {arXiv preprint arXiv:2307.09288}
}

@inproceedings{
hoogeboom2021argmax,
title={Argmax Flows and Multinomial Diffusion: Learning Categorical Distributions},
author={Emiel Hoogeboom and Didrik Nielsen and Priyank Jaini and Patrick Forr{\'e} and Max Welling},
booktitle={Advances in Neural Information Processing Systems},
editor={A. Beygelzimer and Y. Dauphin and P. Liang and J. Wortman Vaughan},
year={2021},
url={https://openreview.net/forum?id=6nbpPqUCIi7}
}

@article{austin2021structured,
  title={Structured denoising diffusion models in discrete state-spaces},
  author={Austin, Jacob and Johnson, Daniel D and Ho, Jonathan and Tarlow, Daniel and Van Den Berg, Rianne},
  journal={Advances in neural information processing systems},
  volume={34},
  pages={17981--17993},
  year={2021}
}

@inproceedings{kingma2016improved,
    title = {Improved variational inference with inverse autoregressive flow},
    volume = {29},
    url = {https://proceedings.neurips.cc/paper_files/paper/2016/file/ddeebdeefdb7e7e7a697e1c3e3d8ef54-Paper.pdf},
    booktitle = {Advances in neural information processing systems},
    publisher = {Curran Associates, Inc.},
    author = {Kingma, Durk P and Salimans, Tim and Jozefowicz, Rafal and Chen, Xi and Sutskever, Ilya and Welling, Max},
    editor = {Lee, D. and Sugiyama, M. and Luxburg, U. and Guyon, I. and Garnett, R.},
    year = {2016},
}

@inproceedings{draxler2024freeform,
    title = {Free-form {Flows}: {Make} {Any} {Architecture} a {Normalizing} {Flow}},
    booktitle = {Artificial {Intelligence} and {Statistics}},
    author = {Draxler, Felix and Sorrenson, Peter and Zimmermann, Lea and Rousselot, Armand and Köthe, Ullrich},
    year = {2024},
}

@inproceedings{xue2024easytpp,
    title = {{EasyTPP}: {Towards} open benchmarking temporal point processes},
    url = {https://arxiv.org/abs/2307.08097},
    booktitle = {International conference on learning representations ({ICLR})},
    author = {Xue, Siqiao and Shi, Xiaoming and Chu, Zhixuan and Wang, Yan and Hao, Hongyan and Zhou, Fan and Jiang, Caigao and Pan, Chen and Zhang, James Y. and Wen, Qingsong and Zhou, Jun and Mei, Hongyuan},
    year = {2024},
}

@misc{khoshnoodi2024comprehensive,
    title = {A {Comprehensive} {Survey} of {Accelerated} {Generation} {Techniques} in {Large} {Language} {Models}},
    url = {http://arxiv.org/abs/2405.13019},
    doi = {10.48550/arXiv.2405.13019},
    urldate = {2025-09-25},
    publisher = {arXiv},
    author = {Khoshnoodi, Mahsa and Jain, Vinija and Gao, Mingye and Srikanth, Malavika and Chadha, Aman},
    month = may,
    year = {2024},
    note = {arXiv:2405.13019 [cs]},
}

@inproceedings{paszke2019pytorch,
  title = {Pytorch: {{An}} Imperative Style, High-Performance Deep Learning Library},
  booktitle = {Advances in {{Neural Information Processing Systems}}},
  author = {Paszke, Adam and Gross, Sam and Massa, Francisco and Lerer, Adam and Bradbury, James and Chanan, Gregory and Killeen, Trevor and Lin, Zeming and Gimelshein, Natalia and Antiga, Luca and others},
  year = {2019}
}

@article{harris2020array,
  title = {Array Programming with {{NumPy}}},
  author = {Harris, Charles R. and Millman, K. Jarrod and {van der Walt}, St{\'e}fan J. and Gommers, Ralf and Virtanen, Pauli and Cournapeau, David and Wieser, Eric and Taylor, Julian and Berg, Sebastian and Smith, Nathaniel J. and Kern, Robert and Picus, Matti and Hoyer, Stephan and {van Kerkwijk}, Marten H. and Brett, Matthew and Haldane, Allan and {del R{\'i}o}, Jaime Fern{\'a}ndez and Wiebe, Mark and Peterson, Pearu and {G{\'e}rard-Marchant}, Pierre and Sheppard, Kevin and Reddy, Tyler and Weckesser, Warren and Abbasi, Hameer and Gohlke, Christoph and Oliphant, Travis E.},
  year = {2020},
  journal = {Nature},
  volume = {585},
  number = {7825},
  pages = {357--362}
}

@inproceedings{mckinney2010data,
  title = {Data {{Structures}} for {{Statistical Computing}} in {{Python}}},
  booktitle = {9th {{Python}} in {{Science Conference}}},
  author = {McKinney, Wes},
  editor = {{van der Walt}, St{\'e}fan and {Jarrod Millman}},
  year = {2010}
}

@misc{falcon2019pytorch,
  title = {{{PyTorch}} Lightning},
  author = {Falcon, William and {The PyTorch Lightning team}},
  year = {2019},
  month = mar,
  copyright = {Apache-2.0},
  version = {1.4}
}

@article{hunter2007matplotlib,
  title = {Matplotlib: {{A 2D}} Graphics Environment},
  author = {Hunter, J. D.},
  year = {2007},
  journal = {Computing in Science \& Engineering},
  volume = {9},
  number = {3},
  pages = {90--95}
}

@article{hinton2015distilling,
  title={Distilling the knowledge in a neural network},
  author={Hinton, Geoffrey and Vinyals, Oriol and Dean, Jeff},
  journal={arXiv preprint arXiv:1503.02531},
  year={2015}
}

@inproceedings{
draxler2025transformers,
title={Transformers for Mixed-type Event Sequences},
author={Felix Draxler and Yang Meng and Kai Nelson and Lukas Laskowski and Yibo Yang and Theofanis Karaletsos and Stephan Mandt},
booktitle={The Thirty-ninth Annual Conference on Neural Information Processing Systems},
year={2025},
url={https://openreview.net/forum?id=MtwsRjPZhf}
}

@inproceedings{
liu2025discrete,
title={Discrete Copula Diffusion},
author={Anji Liu and Oliver Broadrick and Mathias Niepert and Guy Van den Broeck},
booktitle={The Thirteenth International Conference on Learning Representations},
year={2025},
url={https://openreview.net/forum?id=FXw0okNcOb}
}

@inproceedings{
feng2025theoretical,
title={Theoretical Benefit and Limitation of Diffusion Language Model},
author={Guhao Feng and Yihan Geng and Jian Guan and Wei Wu and Liwei Wang and Di He},
booktitle={The Thirty-ninth Annual Conference on Neural Information Processing Systems},
year={2025},
url={https://openreview.net/forum?id=fGBCRZQVse}
}

@inproceedings{oord2018parallel,
  title={Parallel wavenet: Fast high-fidelity speech synthesis},
  author={Oord, Aaron and Li, Yazhe and Babuschkin, Igor and Simonyan, Karen and Vinyals, Oriol and Kavukcuoglu, Koray and Driessche, George and Lockhart, Edward and Cobo, Luis and Stimberg, Florian and others},
  booktitle={International conference on machine learning},
  pages={3918--3926},
  year={2018},
  organization={PMLR}
}

@article{samragh2025your,
  title={Your llm knows the future: Uncovering its multi-token prediction potential},
  author={Samragh, Mohammad and Kundu, Arnav and Harrison, David and Nishu, Kumari and Naik, Devang and Cho, Minsik and Farajtabar, Mehrdad},
  journal={arXiv preprint arXiv:2507.11851},
  year={2025}
}

@article{li2024eagle1,
  title={Eagle: Speculative sampling requires rethinking feature uncertainty},
  author={Li, Yuhui and Wei, Fangyun and Zhang, Chao and Zhang, Hongyang},
  journal={arXiv preprint arXiv:2401.15077},
  year={2024}
}

@article{li2024eagle,
  title={Eagle-2: Faster inference of language models with dynamic draft trees},
  author={Li, Yuhui and Wei, Fangyun and Zhang, Chao and Zhang, Hongyang},
  journal={arXiv preprint arXiv:2406.16858},
  year={2024}
}

@inproceedings{hu2025sam,
  title={Sam decoding: Speculative decoding via suffix automaton},
  author={Hu, Yuxuan and Wang, Ke and Zhang, Xiaokang and Zhang, Fanjin and Li, Cuiping and Chen, Hong and Zhang, Jing},
  booktitle={Proceedings of the 63rd Annual Meeting of the Association for Computational Linguistics (Volume 1: Long Papers)},
  pages={12187--12204},
  year={2025}
}

@article{ankner2024hydra,
  title={Hydra: Sequentially-dependent draft heads for medusa decoding},
  author={Ankner, Zachary and Parthasarathy, Rishab and Nrusimha, Aniruddha and Rinard, Christopher and Ragan-Kelley, Jonathan and Brandon, William},
  journal={arXiv preprint arXiv:2402.05109},
  year={2024}
}

@article{cai2024medusa,
  title={Medusa: Simple llm inference acceleration framework with multiple decoding heads},
  author={Cai, Tianle and Li, Yuhong and Geng, Zhengyang and Peng, Hongwu and Lee, Jason D and Chen, Deming and Dao, Tri},
  journal={arXiv preprint arXiv:2401.10774},
  year={2024}
}

@inproceedings{luo2025turning,
  title={Turning trash into treasure: Accelerating inference of large language models with token recycling},
  author={Luo, Xianzhen and Wang, Yixuan and Zhu, Qingfu and Zhang, Zhiming and Zhang, Xuanyu and Yang, Qing and Xu, Dongliang},
  booktitle={Proceedings of the 63rd Annual Meeting of the Association for Computational Linguistics (Volume 1: Long Papers)},
  pages={6816--6831},
  year={2025}
}

@article{fu2024break,
  title={Break the sequential dependency of llm inference using lookahead decoding},
  author={Fu, Yichao and Bailis, Peter and Stoica, Ion and Zhang, Hao},
  journal={arXiv preprint arXiv:2402.02057},
  year={2024}
}

@misc{saxena2023prompt,
    title = {Prompt Lookup Decoding},
    author = {Apoorv Saxena},
    year = {2023},
    month = {November},
    url = {https://github.com/apoorvumang/prompt-lookup-decoding/}
}

@misc{vicuna2023,
    title = {Vicuna: An Open-Source Chatbot Impressing GPT-4 with 90\%* ChatGPT Quality},
    url = {https://lmsys.org/blog/2023-03-30-vicuna/},
    author = {Chiang, Wei-Lin and Li, Zhuohan and Lin, Zi and Sheng, Ying and Wu, Zhanghao and Zhang, Hao and Zheng, Lianmin and Zhuang, Siyuan and Zhuang, Yonghao and Gonzalez, Joseph E. and Stoica, Ion and Xing, Eric P.},
    month = {March},
    year = {2023}
}

@inproceedings{chen2024sharegpt4v,
  title={Sharegpt4v: Improving large multi-modal models with better captions},
  author={Chen, Lin and Li, Jinsong and Dong, Xiaoyi and Zhang, Pan and He, Conghui and Wang, Jiaqi and Zhao, Feng and Lin, Dahua},
  booktitle={European Conference on Computer Vision},
  pages={370--387},
  year={2024},
  organization={Springer}
}

@article{xia2024unlocking,
  title={Unlocking efficiency in large language model inference: A comprehensive survey of speculative decoding},
  author={Xia, Heming and Yang, Zhe and Dong, Qingxiu and Wang, Peiyi and Li, Yongqi and Ge, Tao and Liu, Tianyu and Li, Wenjie and Sui, Zhifang},
  journal={arXiv preprint arXiv:2401.07851},
  year={2024}
}

@article{hu2022lora,
  title={Lora: Low-rank adaptation of large language models.},
  author={Hu, Edward J and Shen, Yelong and Wallis, Phillip and Allen-Zhu, Zeyuan and Li, Yuanzhi and Wang, Shean and Wang, Lu and Chen, Weizhu and others},
  journal={ICLR},
  volume={1},
  number={2},
  pages={3},
  year={2022}
}

@article{wolf2019huggingface,
  title={Huggingface's transformers: State-of-the-art natural language processing},
  author={Wolf, Thomas and Debut, Lysandre and Sanh, Victor and Chaumond, Julien and Delangue, Clement and Moi, Anthony and Cistac, Pierric and Rault, Tim and Louf, R{\'e}mi and Funtowicz, Morgan and others},
  journal={arXiv preprint arXiv:1910.03771},
  year={2019}
}
